%% file: meta_arxiv.tex
\PassOptionsToPackage{table,dvipsnames}{xcolor}
\documentclass{fairmeta}
\usepackage[utf8]{inputenc}
\input{preamble}

\title{\ours: Logically Think with Images for Visual Search}

\author[1,2,*]{Yanjun Fu}
\author[1,\dagger]{Quanzeng You}
\author[1]{Jiadong Guo}
\author[1]{Yujie Lu}
\author[1]{Sai Charitha Akula}
\author[1]{Haiping Wu}
\author[1,\dagger]{Miao Liu}
\author[2,\dagger]{Sanghamitra Dutta}
\author[1,\S]{Lu Yuan}

\affiliation[1]{Meta Superintelligence Labs (MSL)}
\affiliation[2]{University of Maryland}
\contribution[*]{Work was done at Meta}
\contribution[\dagger]{Equal advising}
\contribution[\S]{Project lead}

\input{sec/0_abstract}
\correspondence{\email{yanjunfu@umd.edu}, \email{qyou@meta.com}}

\hypersetup{
  pdftitle={LogT: Logically Think with Images for Visual Search},
  pdfauthor={Yanjun Fu, Quanzeng You, Jiadong Guo, Yujie Lu, Sai Charitha Akula, Haiping Wu, Miao Liu, Sanghamitra Dutta, Lu Yuan},
  pdfsubject={Visual search with multimodal language models}
}

\begin{document}
\maketitle

\input{sec/1_intro}
\input{sec/2_related}
\input{sec/3_approach}
\input{sec/4_experiments}
\input{sec/5_discussion}
\input{sec/6_conclusion}
\input{sec/ack}

\clearpage
\bibliographystyle{assets/arxivnat}
\bibliography{main}

\input{sec/appendix}

\end{document}

%% file: preamble.tex
\makeatletter \DeclareRobustCommand\onedot{\futurelet\@let@token\@onedot} \def\@onedot{\ifx\@let@token.\else.\null\fi\xspace} \def\eg{\emph{e.g}\onedot}  \def\ie{\emph{i.e}\onedot}    \def\etc{\emph{etc}\onedot}      \def\etal{\emph{et al}\onedot} \makeatother

\usepackage{enumitem}
\usepackage{xspace}
\usepackage{float}
\usepackage{wrapfig}
\usepackage{multirow}
\usepackage{pifont}
\usepackage[ruled,vlined]{algorithm2e}
\usepackage{tablefootnote}
\usepackage{adjustbox}
\usepackage{subcaption}

\newcommand{\myparagraph}[1]{\textbf{#1}\hspace{0.25em}}

\usepackage[most]{tcolorbox}
\tcbuselibrary{listings}

\definecolor{qagenblue}{RGB}{218, 237, 253}
\definecolor{tracegenpurple}{RGB}{234, 223, 253}
\definecolor{qagendarkblue}{RGB}{124, 190, 249}
\definecolor{tracegendarkpurple}{RGB}{182, 136, 248}

\newtcolorbox{qagenbox}[1]{
  enhanced,          
  unbreakable,   
  colback=qagenblue!20,
  colframe=qagendarkblue,
  colbacktitle=qagendarkblue,
  coltitle=white,
  fonttitle=\bfseries,
  boxrule=2pt,
  arc=2mm,
  auto outer arc,
  left=1pt,
  right=1pt,
  top=1pt,
  bottom=1pt,
  title=#1
}

\newtcolorbox{tracegenbox}[1]{
  enhanced,          
  unbreakable,       
  colback=tracegenpurple!25,
  colframe=tracegendarkpurple,
  colbacktitle=tracegendarkpurple,
  coltitle=white,
  fonttitle=\bfseries,
  boxrule=2pt,
  arc=2mm,
  auto outer arc,
  left=1pt,
  right=1pt,
  top=1pt,
  bottom=1pt,
  title=#1
}

\newtcolorbox{humanevalbox}[1]{
  enhanced,          
  unbreakable,       
  colback=gray!10,
  colframe=gray!80,
  colbacktitle=gray!80,
  coltitle=white,
  fonttitle=\bfseries,
  boxrule=2pt,
  arc=2mm,
  auto outer arc,
  left=1pt,
  right=1pt,
  top=1pt,
  bottom=1pt,
  title=#1
}

\definecolor{myred}{RGB}{247, 194, 190}

\newcommand{\ours}{\textsc{LogT}\xspace}

\newcommand{\zoom}{\textsc{Zoom-in}\xspace}

\newcommand{\qwenvl}{Qwen2.5-VL-7B\xspace}
\newcommand{\qwenvlnew}{Qwen3-VL-8B\xspace}
\newcommand{\qwenvlmid}{Qwen2.5-VL-32B\xspace}
\newcommand{\qwenvlxl}{Qwen2.5-VL-72B\xspace}
\newcommand{\llavaov}{LLaVA-OneVision-7B\xspace}
\newcommand{\internvl}{InternVL3.5-8B\xspace}
\newcommand{\gpt}{GPT-OSS-120B\xspace}
\newcommand{\gptsmall}{GPT-OSS-20B\xspace}

\newcommand{\sketch}{\textsc{Visual Sketchpad}\xspace}
\newcommand{\pyvision}{\textsc{PyVision}\xspace}

\newcommand{\seal}{\textsc{SEAL}\xspace}
\newcommand{\dyfo}{\textsc{DyFo}\xspace}

\newcommand{\pixel}{\textsc{Pixel Reasoner}\xspace}
\newcommand{\cof}{\textsc{Chain-of-Focus}\xspace}
\newcommand{\deep}{\textsc{DeepEyes}\xspace}
\newcommand{\minio}{\textsc{Mini-o3}\xspace}

\newcommand{\avg}{average@8\xspace}
\newcommand{\pass}{pass@1\xspace}

\newcommand{\seg}{SA1B\xspace}
\newcommand{\vstar}{V\textsuperscript{*}\xspace}
\newcommand{\visualp}{VisualProbe\xspace}
\newcommand{\hrbench}{HR-Bench\xspace}
\newcommand{\mmelong}{MME-RealWorld-Lite\xspace}
\newcommand{\mme}{MME-RW-Lite\xspace}

\newcommand{\cmark}{\ding{51}}
\newcommand{\xmark}{\ding{55}}
\newcommand{\csymbol}{\cmark\xspace}
\newcommand{\xsymbol}{\xmark\xspace}
\colorlet{darkgreen}{green!65!black}
\colorlet{darkred}{red!80!black}

\newcommand{\Inc}[1]{\textcolor{darkgreen}{\textbf{\small{+#1}}}}
\newcommand{\Dec}[1]{\textcolor{darkred}{\textbf{\small{-#1}}}}

\DeclareFontShape{T1}{optimistic}{b}{sc}{<-> s * [0.88] ./assets/optimistic}{}
\crefname{figure}{Figure}{Figures}
\crefname{table}{Table}{Tables}
\crefname{section}{Section}{Sections}
\crefname{equation}{Equation}{Equations}

%% file: sec/0_abstract.tex
\abstract{
Thinking with images empowers vision-language models (VLMs) to use an external \zoom tool for fine-grained visual search.
Recent works employ supervised fine-tuning (SFT) and reinforcement learning (RL) with tool-specific designs to enable such capabilities.
However, achieving logical reasoning alongside accurate and efficient tool use remains a significant challenge.
To address this, we shift the focus away from increasing RL complexity and instead identify SFT data quality as a key bottleneck.
We then propose \ours (short for \textbf{Log}ically \textbf{T}hink with Images), a fully automated SFT data engine that explicitly enforces logic-chain-guided reasoning.
\ours generates an SFT dataset featuring challenging, visually grounded questions and structured tool-use traces with human-like logic.
We validate our approach by fine-tuning two base models, \qwenvl and \internvl, on the dataset and evaluating them across seven benchmark test sets.
We show that \ours-SFT based on \qwenvl outperforms the state-of-the-art trained with both SFT and RL, despite using 3.1$\times$ less data and $>$120$\times$ fewer GPU hours. 
Furthermore, applying a simple RL recipe without tool-specific design on \ours-SFT yields a model that surpasses prior methods with complex RL setups by a substantial 9.59-point margin.
In addition to strong performance gains, our model exhibits higher tool-use accuracy and efficiency, as well as more structured and logically consistent reasoning traces, directly validating the effectiveness of our approach.
}

%% file: sec/1_intro.tex
\section{Introduction}
\label{sec:intro}
Vision-Language Models (VLMs)~\citep{alayrac2022flamingo,liu2023visual} integrate visual and textual information by aligning high-level visual cues with text embeddings via visual encoders and projection layers.
However, this alignment process introduces irreversible information loss, causing VLMs to struggle with visually intensive tasks that require fine-grained visual search~\citep{wu2023vstar}.
Even when augmented with text-based chain-of-thought (CoT) reasoning~\citep{wei2022cot,zhang2025improve}, such limitations persist, as textual reasoning cannot recover missing visual details.
Therefore, recent works~\citep{hu2024visual,openaio32025} introduce thinking with images, enabling VLMs to invoke external tools to extract additional visual information and interleave it with textual reasoning.
In this context, we investigate how VLMs can more effectively leverage the widely studied \zoom tool~\citep{openaio32025, su2025pixel, zhang2025chain, zheng2025deepeyes, lai2025mini, li2025looklessreasonmore} for visual search.

Existing efforts approach this problem through post-training, typically combining supervised fine-tuning (SFT) for cold start with reinforcement learning (RL) driven by outcome-based and tool-related rewards.
While these methods have shown promising results, we find that the improvement largely stems from increased familiarity with the tool format gained during post-training.
In contrast, enabling VLMs to logically reason about when and how to invoke the tool for accurate region-of-interest identification and efficient inference remains a fundamental challenge~\citep{openai2025thinking} (see analysis in~\cref{sec:analysis}).
While RL with finer-grained step-by-step process supervision using process reward models (PRMs)~\citep{li2026toolrm} has shown early promise in improving reasoning aided by non-visual tools, PRMs for training VLMs to think with images and use \zoom effectively and efficiently remain underexplored, largely due to the limited scalability and high compute cost of training dedicated PRMs.

Therefore, we shift the focus from designing more complex RL to improving SFT data quality, which we identify as \emph{a key bottleneck}.
We propose \ours, a novel data engine that generates SFT data to teach VLMs to \textbf{Log}ically \textbf{T}hink with Images for visual search. Importantly, \ours is fully automated and relies solely on open-source models to enhance the quality of SFT data.
Each synthesized SFT sample includes a visual search question-answer (QA) pair, and the corresponding tool-use reasoning traces.
Accordingly, we design \ours to improve the quality of both QAs and reasoning traces.

\input{figures/teaser}
For QA quality, we enrich the VLM’s input with detailed visual contexts in addition to the full image, enabling the synthesis of more challenging questions that target fine-grained visual details. We then apply rigorous filtering to retain only high-quality QA pairs.
For reasoning trace quality, we first construct logic chains that mimic human problem-solving strategies (\eg, multi-hop reasoning in~\cref{fig:main_idea}).
These logic chains serve as skeletons and outline the key steps needed to solve each visual search task. We then use them to guide the VLM in generating structured and coherent reasoning traces with appropriate tool calls.

To demonstrate the effectiveness of \ours, we train \qwenvl~\citep{bai2025qwen2} and \internvl~\citep{wang2025internvl3_5} on our synthesized SFT dataset comprising 7.1k high-quality entries.
As shown in \cref{fig:result_highlights}, using 3.1$\times$ less data and $>$120$\times$ fewer GPU hours, our SFT-only model (\ours-SFT based on \qwenvl) outperforms the state-of-the-art (SOTA) open-source model trained by both SFT and RL across seven test sets.
We further apply off-the-shelf RL with straightforward outcome-based rewards, and the resulting model outperforms prior methods by at least 9.59 points, despite their reliance on more complex, tool-specific RL strategies.
In addition, both our SFT and RL checkpoints achieve better tool-call accuracy and efficiency. A human evaluation further shows that our method produces more coherent and higher-quality reasoning traces.

In summary, our contributions are threefold:
\begin{itemize}[leftmargin=2em, topsep=0pt, parsep=0pt]
    \item We identify SFT data quality as a key bottleneck and introduce \ours, a novel data synthesis strategy that generates high-quality SFT data with challenging QA pairs and coherent, human-like reasoning traces.
    This strategy produces a dataset with 7.1k entries.
    \item We show that training \qwenvl solely on our SFT data outperforms prior SOTA models that use both SFT and RL, while also achieving significantly higher data and training efficiency, highlighting the benefit of investing more compute and effort in SFT data quality.
    \item We further improve \ours-SFT with a simple RL setup without any tool-specific design, underscoring the importance of SFT quality.
    The resulting model also achieves higher tool-call accuracy, greater tool-call efficiency, and more logically coherent reasoning traces.
\end{itemize}

%% file: figures/teaser.tex
\begin{figure}[t]
    \centering
    \begin{minipage}{0.48\linewidth}
        \centering
        \includegraphics[width=0.95\linewidth]{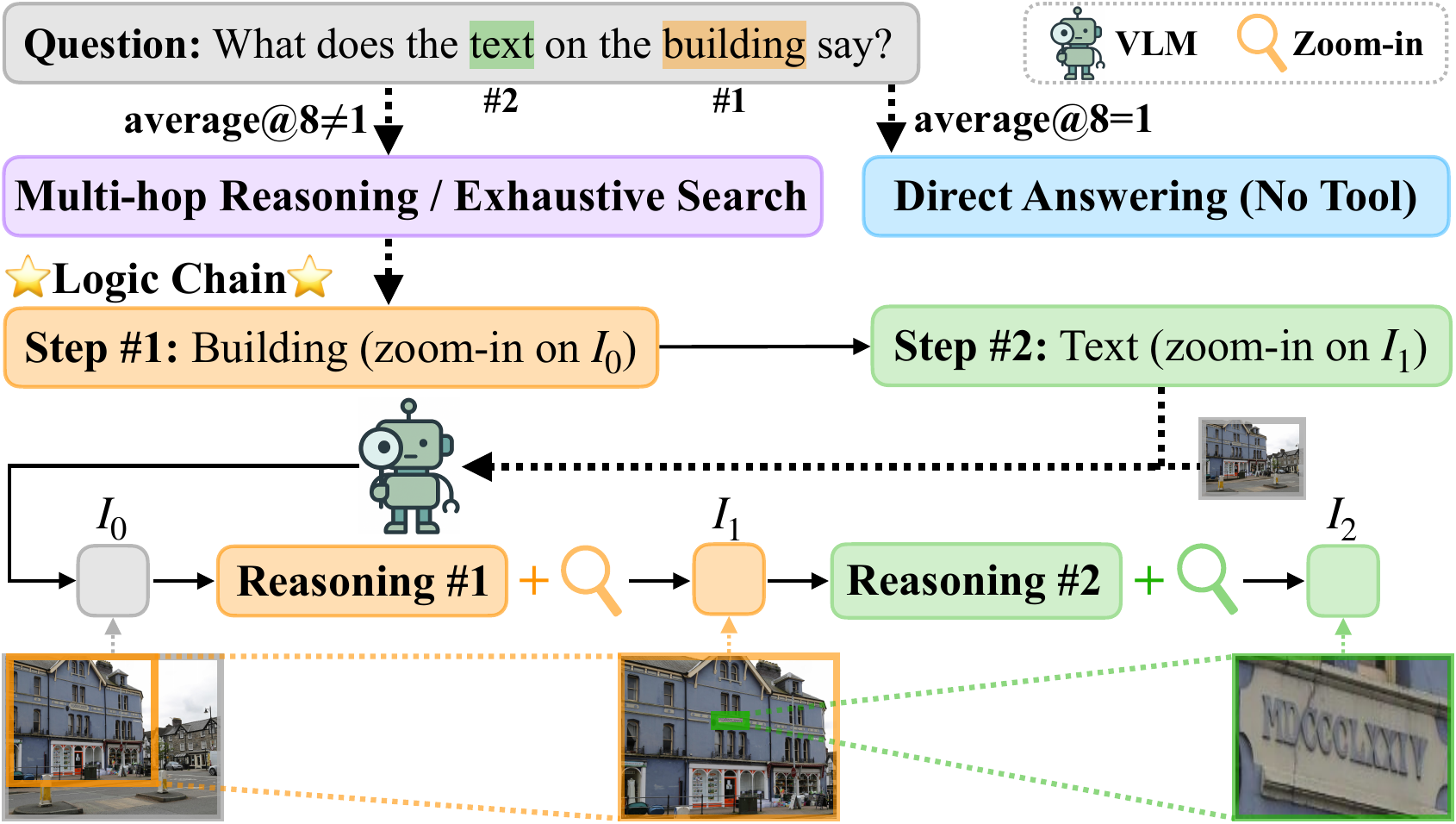}
        \caption{\textbf{Main idea.} We generate reasoning traces based on human-like strategies including~(i) \emph{direct answering} for simple questions,~(ii)~\emph{multi-hop reasoning} by focusing on different regions, and~(iii) \emph{exhaustive search} for more difficult cases. For easy questions where the base model's \avg is 1, we synthesize direct answering traces without \zoom. Otherwise, we first generate \emph{logic chains} that specify key tool-use steps and then use them to guide the VLM in synthesizing tool-use reasoning traces.}
        \label{fig:main_idea}
    \end{minipage}
    \hfill
    \begin{minipage}{0.48\linewidth}
        \centering
        \includegraphics[width=0.95\linewidth]{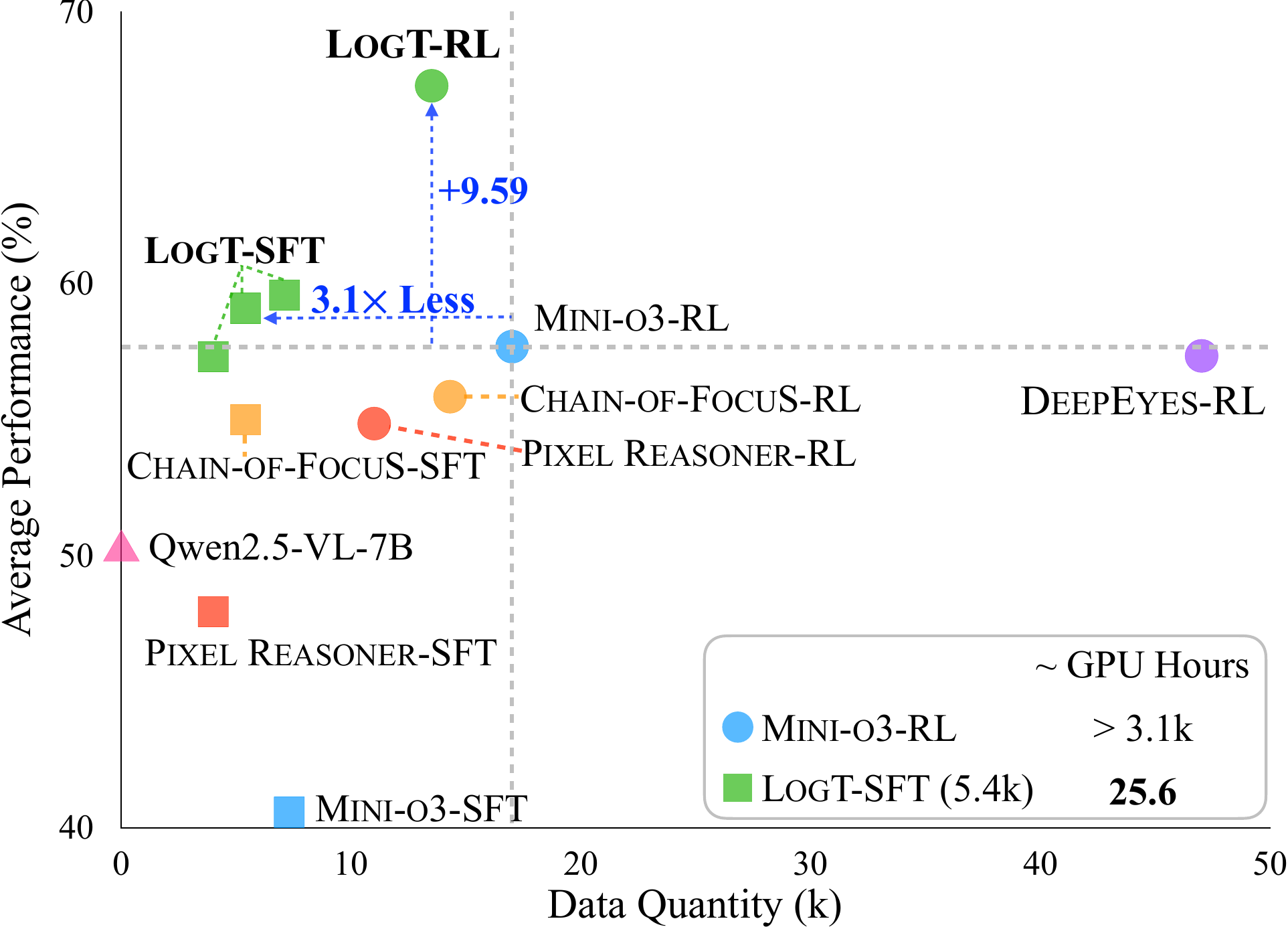}
        \caption{\textbf{Result highlights.} By applying SFT to a \qwenvl model using 5.4k to 7.1k of our synthesized data, \ours-SFT surpasses the best-performing \minio-RL on average across seven test sets, while using up to 3.1× less data and $>$120$\times$ fewer GPU hours. With a simple RL setup without any tool-specific design, \ours-RL further outperforms baselines by 9.59.}
        \label{fig:result_highlights}
    \end{minipage}
    \vspace{-0.5em}
\end{figure}

%% file: sec/2_related.tex
\section{Related Work}
\label{sec:related}

\myparagraph{Think with images.}
Thinking with images extends text-only CoT reasoning by allowing VLMs to interleave textual reasoning with visual information extracted via external tools from either the input image or intermediate tool outputs.
Some methods \citep{hu2024visual, zhao2025pyvision} adopt a training-free, modular agent-based design, allowing VLMs (\eg, GPT-4o \citep{hurst2024gpt}) to invoke visual tools during inference.
In contrast, other approaches, including OpenAI's o3 \citep{openaio32025, huang2025vista, huang2025samr1, wu2025vtoolr1}, rely on post-training to incentivize VLMs with tool-use capabilities.

Among prior work, \zoom is one of the most widely used visual tools~\citep{openaio32025, su2025pixel, zhang2025chain, zheng2025deepeyes, lai2025mini, li2025looklessreasonmore, su2025openthinkimg, zhang2025thyme, liu2025faithfulness}, showing consistent improvements in visual search~\citep{wu2023vstar}.
Most methods employ a two-stage post-training pipeline of SFT followed by RL.
SFT primarily teaches models to invoke \zoom and conform to its calling format, which is useful because current VLMs exhibit limited zero-shot tool-use ability.
RL then uses outcome-based rewards to further encourage tool usage and format compliance.
However, these approaches still face fundamental challenges to train VLMs to reason logically with images and to call \zoom both accurately~\citep{liu2025faithfulness} and efficiently~\citep{openai2025thinking}.
Li~\etal \citep{li2025looklessreasonmore} improve tool-call efficiency through designing complex RL rewards and rollout strategies, but at the tradeoff of substantial RL compute overhead.
RL with process supervision could potentially mitigate these limitations.
However, this often requires PRMs~\citep{li2026toolrm}.
To the best of our knowledge, PRMs tailored for thinking with images remain underexplored, and training such specialized PRMs that meet our requirements is both challenging and very intensive.

\myparagraph{SFT data for visual reasoning.}
Building SFT datasets for thinking with images requires reasoning traces that interleave text with images returned by tool calls.
This is challenging because most existing SFT datasets for visual reasoning~\citep{lu2022learn, mu2023embodied, xu2025llavacotletvisionlanguage, guo_2025_mammoth, sun_2025_mmverify} contain only textual reasoning, without intermediate visual evidence.
A few datasets~\citep{shao2024visualcot, li2025zebracot} do provide reasoning traces that interleave texts and images using referenced bounding boxes.
However, they typically use low-resolution images and include only a single bounding box per trace, making them inadequate for high-resolution and fine-grained visual search tasks requiring multi-step tool use.

Recent works on thinking with images~\citep{su2025pixel,zhang2025chain, lai2025mini, zhang2025thyme} release their own SFT datasets.
These studies typically generate visual search questions by directly prompting VLMs with full input images.
Consequently, the synthesized questions often center on easy-to-search objects.
Furthermore, the reasoning traces are usually generated by directly prompting VLMs or agents without careful quality control, and they often lack the human-like logical structure for solving complex problems, as presented in our approach.
We also note that, despite the release of datasets and checkpoints in prior work, the data generation engines themselves receive less attention. In contrast, we focus on the data generation strategy itself.

%% file: sec/3_approach.tex
\section{Our Data Engine: \ours}
\label{sec:logt}
We propose a novel data synthesis framework, \ours, for generating an SFT dataset with high-quality visual search QA pairs and reasoning traces.
This dataset is used to enhance VLMs' ability to use \zoom for fine-grained visual search.
Our goal is to train VLMs not only to invoke \zoom in the correct format over one or multiple turns, but also to logically think with images and to use \zoom accurately and efficiently to identify regions of interest.

\cref{fig:logt_overview} illustrates the process for generating a single-object QA and its corresponding tool-use reasoning trace.
Given an image $I_0$ and two object-centric crops that provide contextual ($I_C$) and focused ($I_F$) views of the target object, the QA generator produces candidate QA pairs, which are then filtered using automatic quality checks.
We then pass the generated QAs along with a synthesized caption of $I_0$ to a logic chain generator, which produces a multi-hop chain $L_M$ and an exhaustive search chain ($L_E$). One of these chains is selected and used to guide the reasoning trace generator to construct a full trace that interleaves textual reasoning with \zoom actions. Each \zoom call extracts a region from $I_0$ or a previous crop, resulting in crops denoted as $I_1$ through $I_3$.

While ensuring high data quality, \ours is \emph{fully automated} and \emph{accessible}, relying solely on open-source models without needing complex visual agents. In the following sections, we detail the key components of (i) QA pair generation (\cref{sec:qa_synthesis}) and (ii) reasoning trace synthesis (\cref{sec:trace_synthesis}).

\input{figures/logt_overview_figure}
\subsection{Object-oriented QA Pair Synthesis.}
\label{sec:qa_synthesis}
Our QA pair synthesis strategy consists of two steps:~(i) generating challenging QA pairs using a QA generator,~and (ii) performing quality control through a filter and a QA quality judge.
Specifically, we use \qwenvlxl~\citep{bai2025qwen2} as the generator and a different open-source VLM with 106B parameters as the judge.

As preparation, we source high-resolution images from the \seg dataset~\citep{kirillov2023segment}, similar to \pixel~\citep{su2025pixel} and \cof~\citep{zhang2025chain}. 
For each image, \seg provides bounding boxes representing unnamed objects.

\myparagraph{Enhance QA generation with object-oriented visual contexts.}
We generate visual search QA pairs in an \emph{object-oriented} manner, with each focusing on one or two specific objects within the input image $I_0$.
To avoid bias toward easily identifiable objects and increase question difficulty, we first filter out bounding boxes that cover more than 1\% of the image area.
From the remaining boxes, we randomly select one box for single-object questions, or a pair of boxes for relation-based questions.
In addition to the full image, we prepare two more crops related to each bounding box:~(i) a contextual crop $I_C$ that is several times larger than the bounding box to capture surrounding visual context, and~(ii) a focused crop $I_F$ that exactly matches the bounding box to capture fine-grained object details.
This results in 3 images for single-object and 5 for relation-based questions.
We provide the QA generator with these images, detailed generation rules, a taxonomy of permissible question types, and in-context examples to produce two candidate QA pairs.
The QA generation prompt is shown in~\cref{fig:qagen_prompt}.

\myparagraph{QA filtering and selection.}
We first apply a simple rule-based filter to eliminate QAs that violate output format requirements.
Then we evaluate the remaining pairs using the QA quality judge to determine whether one or both conform to our quality criteria, including answer correctness, question unambiguity, \etc
If neither QA candidate passes, we discard both; if both are valid, the judge selects the better one.
To mitigate positional bias of the judge~\citep{shi2024judging}, we run the evaluation twice with the QA order reversed.
We err on the side of caution by accepting a QA pair only if it is consistently validated across both evaluations.

Compared to prior work, our method produces more challenging and higher-quality questions.
We define \avg as the accuracy obtained by randomly sampling responses eight times at a temperature of 0.8.
As shown in~\cref{fig:easy_questions}, only 24\% of our synthesized questions are perfectly answered by \qwenvl without tool use, \ie, \avg = 1, substantially fewer than in existing datasets.

\subsection{Logic Chain-driven Reasoning Trace Synthesis.}
\label{sec:trace_synthesis}
In visual QA tasks, humans typically adopt one of the following reasoning strategies. We leverage these strategies to ensure the logical coherence of generated reasoning traces and to minimize unnecessary tool usage:~(i)
\textbf{Direct answering:} When the target object is prominent and the question is straightforward, humans can answer directly without intermediate reasoning.~(ii)
\textbf{Multi-hop reasoning:} For questions requiring indirect visual cues, like ``What color is the backpack worn by the person?'', humans identify intermediate visual anchors (\eg, the person), then refine their focus (\eg, the backpack), reasoning through multiple ``hops''.~(iii)
\textbf{Exhaustive search:} When the target is difficult to locate and the question lacks clear visual references, such as asking to find a person in a crowd, humans may divide the scene into several parts and search systematically.

For easy questions where \qwenvl achieves an \avg of 100\% even without requiring tools, we adopt the direct answering strategy to generate a more tool-efficient reasoning trace, consisting of a single turn without invoking the \zoom tool.

Otherwise, we simulate the remaining two strategies and synthesize reasoning traces that involve one or more tool calls.
This process includes:~(i) abstracting these strategies into \emph{logic chains} using a logic chain generator, where each chain consists of a sequence of key steps that specify where to focus and which image to zoom into,~(ii) selecting the most appropriate logic chain via a logic chain selector, and~(iii) leveraging the selected logic chain as structured guidance for the reasoning trace generator to produce coherent tool-use reasoning traces.
While an ideal system would rely on a single VLM to perform all three parts, current open-sourced VLMs often lack the instruction-following reliability and fine-grained reasoning precision required for high-quality logic chain generation.
To address this, we use \gpt~\citep{openai2025gptoss} as the logic chain generator, and the VLM serves as the QA quality judge for both logic chain selection and reasoning trace generation.

\myparagraph{Logic chain generation and selection.}
Given an image $I_0$ and a QA, we first supply the logic chain generator with the question and a VLM-generated image caption as necessary visual context.
It then outputs two types of logic chains: one for multi-hop reasoning ($L_M$) and one for exhaustive search ($L_E$).
Each chain comprises sequential steps that identify a target object or region (\eg, ``person'' or ``left crowd'') and determine which image to zoom into (\eg, the full image $I_0$ or a crop returned from a previous \zoom call, such as $I_1$ or $I_2$ in~\cref{fig:logt_overview}).
The logic chain generator must reason over spatial and contextual dependencies.
For example, in~\cref{fig:logt_overview}, it needs to decide whether to zoom in gradually, such as identifying the backpack from a crop of the person, or to access multiple objects in parallel, such as finding the security gate and the person.

Once both logic chains are generated, we pass them along with the full image and question to the logic chain selector.
We use it to choose a more appropriate chain based on task complexity and tool call efficiency.

\myparagraph{Reasoning trace synthesis.}
Next, we prompt the reasoning trace generator to follow the selected logic chain step-by-step, producing a reasoning thought and a corresponding \zoom action at each turn until reaching the final answer.
The prompt used for reasoning trace synthesis is shown in~\cref{fig:tracegen_prompt}.
Finally, we build an SFT dataset with 7.1k high-quality QAs and reasoning traces.

We present example multi-hop reasoning traces in~\cref{fig:sft_trace_comparison} and an example exhaustive search reasoning trace in~\cref{fig:exhaustive_search}.
Tool-use statistics of our synthesized SFT dataset are provided in~\cref{appendix:approach_tool_call_distribution} and visualized in~\cref{fig:sft_tool_distribution}.
To qualitatively demonstrate that \ours produces higher-quality reasoning traces, we present a case study comparing our traces with those from prior work in~\cref{appendix:approach_trace_comparision}.
Finally, we analyze the computational cost of our data engine in~\cref{appendix:approach_costs}.

%% file: figures/logt_overview_figure.tex
\begin{figure*}[!t]
    \centering
    \includegraphics[width=\linewidth]{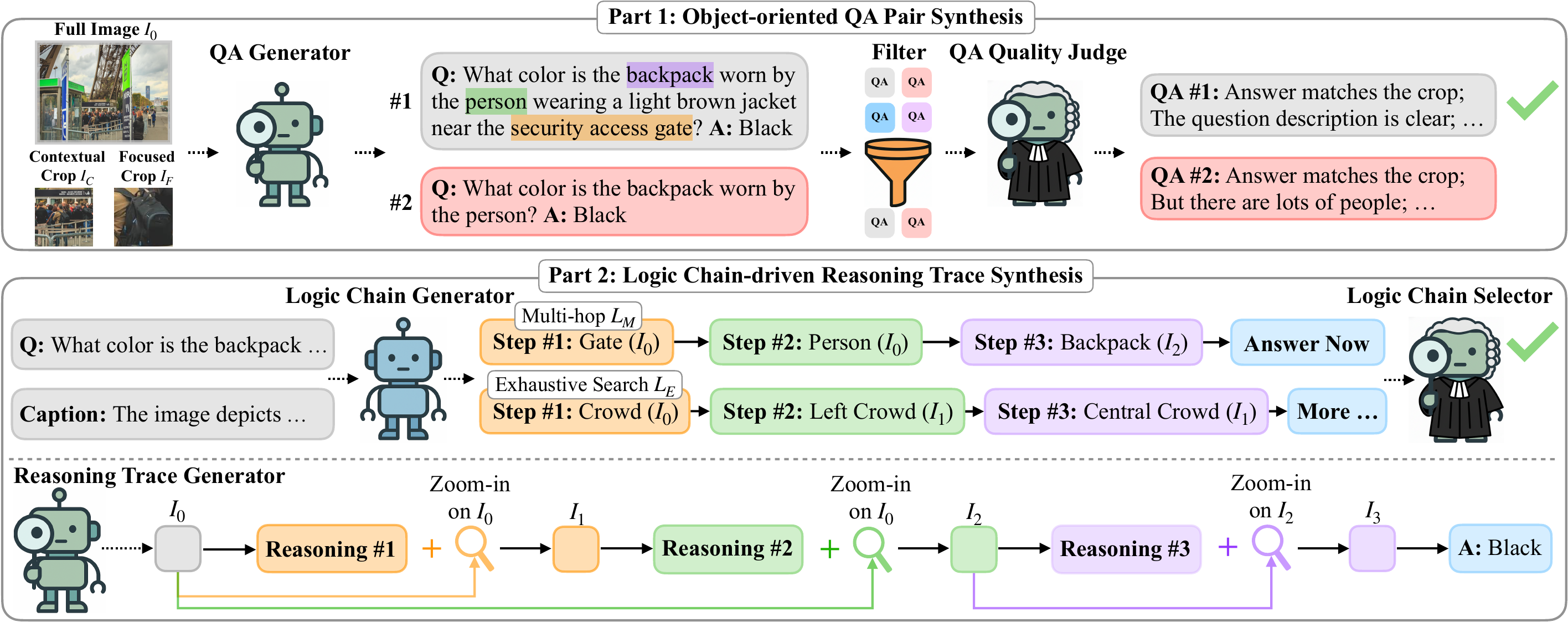}
    \caption{\textbf{A detailed overview of \ours SFT data synthesis engine.} 
    Part 1 (top;~\cref{sec:qa_synthesis}) synthesizes visual search QA pairs using the full image ($I_0$) along with rich visual contexts ($I_C$ and $I_F$, top left) related to randomly selected target objects. After generating candidate QAs, we apply a simple heuristic filter and use a VLM-as-a-Judge to ensure quality. Part 2 (bottom;~\cref{sec:trace_synthesis}) generates reasoning traces for the selected QA. For easy questions, we directly generate traces without tool calls (omitted here). For other questions, we first create logic chains ($L_M$ and $L_E$) based on two human-like strategies. A logic chain selector then chooses the more appropriate one, which is subsequently used to guide the synthesis of multi-turn reasoning traces with tool use.}
    \label{fig:logt_overview}
    \vspace{-0.3em}
\end{figure*}

%% file: sec/4_experiments.tex
\section{Experiments}

\subsection{Experimental Setup}
\label{sec:experiment_setup}
\myparagraph{Baselines.}
We compare our method against a broad set of baselines:~(i) \textbf{VLMs without \zoom}, including GPT-4o~\citep{hurst2024gpt}, and open-source models such as \qwenvl and -72B~\citep{bai2025qwen2}.~(ii)
\textbf{VLMs with \zoom}, including o3~\citep{openaio32025}, and several open-source models post-trained for \zoom, such as \pixel~\citep{su2025pixel}, \deep~\citep{zheng2025deepeyes}, \cof~\citep{zhang2025chain}, \minio~\citep{lai2025mini}, and Li~\etal~\citep{li2025looklessreasonmore}.
All open-source baselines here use \qwenvl as the base model and rely on \zoom as the sole image-interaction tool. They all apply SFT for cold-start training before RL, except for \deep. Notably, all adopt tool-specific RL strategies.
Additional baselines are evaluated in~\cref{tab:other_baselines} and discussed in~\cref{appendix:experiment_baselines}.

\myparagraph{Benchmarks.}
We evaluate our method and baselines on the following benchmarks:~(i) \textbf{Fine-grained visual search on high-resolution images}: This includes \vstar~\citep{wu2023vstar}, \visualp~\citep{lai2025mini}, and \hrbench~\citep{hrbench}.
\visualp is officially divided into three difficulty levels: easy, medium, and hard.
\hrbench consists of two resolution-based subsets: 4K and 8K.~(ii) \textbf{Out-of-domain visual perception and reasoning beyond visual search:} We use \mmelong~\citep{zhang2024mme}, which differs substantially from the \ours SFT dataset in both the visual content and the types of tasks involved.
This makes it well-suited for assessing models' generalization to out-of-domain visual reasoning tasks beyond visual search. \mmelong is a comprehensive benchmark with 1,919 questions covering 43 task types, evaluating general perception and reasoning capabilities of VLMs across diverse scenarios such as OCR, remote sensing, diagram understanding, \etc
In total, our evaluation covers seven test sets, accounting for multiple official splits in \visualp and \hrbench.

\myparagraph{Evaluation details.}
We use greedy decoding for all evaluated models\footnote{\pixel and \cof pad or enlarge model-predicted bounding boxes before applying the \zoom operation. We argue that altering bounding boxes can artificially inflate grounding and tool call accuracy, while increasing inference cost due to additional visual tokens.
To ensure a fair comparison, we disable all bounding-box modifications and evaluate using the model-predicted boxes directly.} and report \pass accuracy to ensure reproducibility.
For all open-sourced models without built-in tool usage, we apply CoT prompting using the instruction ``Please reason step by step.''
For o3, we report the \vstar result from its model card and evaluate \visualp using medium reasoning effort with the code interpreter tool.
Additional evaluation details are provided in~\cref{appendix:experiment_eval_details}.

\input{tables/main_results}

\subsection{Implementation Details}
\label{sec:experiment_implementation}
\myparagraph{SFT implementation.}
Following prior works, we use \qwenvl as the base model.
We train it on the SFT dataset synthesized by \ours for 3 epochs using a learning rate of 1e-5 and an effective batch size of 128. The resulting model is \ours-SFT.

\myparagraph{RL implementation.}
We apply off-the-shelf RL to \ours-SFT fine-tuned on 7.1k samples, to investigate how improved cold-start affects post-RL performance. The resulting model is \ours-RL.
Our reward function consists of only two terms: a correctness reward ($r_{\text{correct}}$) and a formatting reward ($r_{\text{format}}$).
The correctness reward assigns $r_{\text{correct}} = 1$ if the model's predicted answer matches the ground-truth, and $0$ otherwise.
The formatting reward assigns $r_{\text{format}} = 0.5$ if the model follows the correct format, and $0$ otherwise.
The total reward is $r = r_{\text{correct}} + r_{\text{format}}$.

For the RL algorithm, we slightly modify standard Group Relative Policy Optimization (GRPO)~\citep{shao2024deepseekmathpushinglimitsmathematical} to encourage exploration.
Specifically, we remove KL regularization and adopt the clipping-higher strategy recommended by Yu~\etal~\citep{yu2025demystify} for agentic RL to set the upper clipping ratio to 0.315.
We sample 8 rollouts per input, and adopt a constant learning rate of 1e-6.
We use veRL~\citep{sheng2024hybridflow} for implementation and run RL for 100 steps.

We source RL training data from \minio~\citep{lai2025mini}.
Additional details on our RL implementation are provided in~\cref{appendix:experiment_rl_details}.

\subsection{Main SFT Results}
In this section, we report SFT results using \qwenvl as the base model. We also fine-tune \internvl~\citep{wang2025internvl3_5} on our SFT dataset and present those results in~\cref{tab:internvl} of~\cref{appendix:experiment_internvl}.

\myparagraph{\ours-SFT outperforms all baselines, including those trained with both SFT and RL, while being highly data- and training-efficient.}
As shown in~\cref{tab:main_results}, the \ours-SFT model trained on 7.1k synthetic samples generated by our method, surpasses all open-source baselines, including those post-trained with both SFT and RL.  
Compared to the strongest SFT baseline, \cof-SFT, our model achieves higher performance across 6 out of 7 test sets, with an average gain of 4.59 points.
Against the best-performing baseline, \minio-RL, \ours-SFT delivers a 1.9-point improvement.

Moreover, even when the number of training samples is reduced from 7.1k to 5.4k, \ours-SFT still outperforms all baselines. In comparison, \deep-RL uses 8.7$\times$ more data, and \minio-RL uses 3.1$\times$ more.  
Training \qwenvl on 5.4k SFT samples takes only 3.2 hours on 8$\times$A100 GPUs, making it $>$120$\times$ more compute-efficient than the best RL baseline, \minio-RL, in terms of GPU hours. These results underscore the effectiveness of our data engine in producing high-quality SFT training data.

\myparagraph{\ours-SFT significantly outperforms SFT baselines under equal data budgets.}
For fair comparison, we evaluate \ours-SFT under matched or lower data budgets. Specifically, we randomly sample 4k and 5.4k subsets from our synthetic dataset, matching the sizes used by \pixel and \cof, and then fine-tune \qwenvl on these subsets.
Under equal data constraints,~\cref{tab:main_results} shows that \ours-SFT outperforms \pixel-SFT by 9.37 points and \cof-SFT by 4.1 points. With a slightly smaller dataset with 7.1k samples, \ours-SFT surpasses \minio-SFT by a large margin of 19.02 points.  
These results demonstrate the superior quality of our synthetic data generated by our framework.

\input{figures/rl_and_analysis}
\subsection{Main RL Results}
\label{sec:rl_results}
\myparagraph{Applying off-the-shelf RL on top of \ours-SFT improves performance and generalization, outperforming all baselines by a wide margin.}
In~\cref{tab:main_results}, we show that applying off-the-shelf RL to the \ours-SFT model further leads to a notable improvement of 7.69 points.
Remarkably, \ours-RL surpasses all RL baselines by at least 9.59 points, even though those baselines use more sophisticated and tool-specific RL designs.
Moreover, it enhances the model's generalization ability, achieving performance comparable to the current SOTA on the out-of-domain benchmark \mmelong.
These results highlight that the benefits of high-quality SFT data generated by our method are further amplified when combined with RL.

\myparagraph{High-quality SFT data from our framework significantly improves the RL performance ceiling.}
To evaluate the impact of our SFT data engine on the RL performance ceiling, we apply the same RL recipe from~\cref{sec:experiment_implementation} to \qwenvl and other existing SFT baselines~\citep{su2025pixel, zhang2025chain, lai2025mini}.
The validation accuracy dynamics in~\cref{fig:rl_dynamics} show that initializing RL with \ours-SFT leads to much higher performance than applying RL directly to the base model (zero-RL).
In contrast, RL training on prior SFT baselines quickly plateaus after initial gains and often performs no better, or even worse, than zero-RL after 100 steps.
These results indicate that SFT data produced by \ours greatly improve the initial policy and substantially raise the RL performance ceiling. 

\subsection{Analysis on Tool Use and Reasoning Quality}
\label{sec:analysis}
Beyond numerical gains, we show that our data engine enables VLMs to think logically with images and use \zoom accurately and efficiently.
While existing works open-source their training datasets, the SFT data curation pipelines and associated metadata remain unavailable, making apples-to-apples comparisons difficult.
Hence, we focus our analysis on the behavior of the post-trained models. We further show that our model can generalize to novel tools in a zero-shot manner in~\cref{appendix:experiment_rotation}.

\myparagraph{Analysis on tool call accuracy.}
We evaluate the tool accuracy of a model response by assessing how well it localizes the intended region(s) of interest via \zoom.
We first compute the Intersection over Ground-truth (IoG)~\citep{wang18repulsion} for each \zoom call, defined as $\operatorname{IoG} = |B \cap G| / |G|$, where $B$ is the bounding box used in the \zoom call and $G$ is the corresponding ground-truth box.
Next, we consider a response as calling the tool accurately if \emph{at least one} \zoom invocation achieves $\operatorname{IoG} \ge 0.8$, as multiple calls may be issued while answering a question.
\input{figures/tool_efficiency_figure}
Finally, we define a model's overall tool call accuracy as the percentage of tool-requiring responses that include at least one accurate \zoom call.
We compute this metric using \vstar, the only benchmark that provides ground-truth bounding boxes $G$ for target regions.

\cref{fig:tool_accuracy} shows that \ours-SFT achieves 72.09\% tool call accuracy, outperforming both the base model and other SFT baselines. \ours-RL further achieves the highest accuracy of 94.21\%. These results show that the SFT data synthesized by \ours effectively teaches VLMs to reason and invoke tools accurately.
Interestingly, most SFT- and RL-trained baselines underperform the base model that invokes \zoom in a zero-shot manner, except for \minio.
Comparing SFT models and their RL-tuned counterparts, we find that while RL improves tool call accuracy, the quality of SFT is critical for achieving strong tool call accuracy after RL.

\myparagraph{Analysis on tool call efficiency.}
We evaluate tool call efficiency by measuring the average number of \zoom calls used to solve a question across all test sets. As shown in~\cref{fig:tool_efficiency}, the best-performing baseline, \minio-RL, uses an average of 4.3 \zoom calls per task.
In comparison, while delivering higher performance, \ours-SFT and \ours-RL achieve significantly better efficiency, requiring 2.7$\times$ and 2.3$\times$ fewer calls, respectively.
Although other baselines use fewer \zoom calls, their performance is considerably lower.
These results suggest that the SFT data synthesized by \ours enables VLMs to reason about the necessary steps for solving questions, striking a superior balance between tool call efficiency and overall performance.

\myparagraph{Human evaluation on reasoning quality and logical coherence.}
To assess whether the SFT data synthesized by \ours enables VLMs to reason logically with images beyond achieving strong benchmark performance, we conduct a human evaluation comparing \ours-RL with the prior SOTA (\minio-RL) based on overall benchmark scores and tool-use accuracy.
We randomly sample 100 questions from \vstar and ask three human annotators to compare the responses produced by both models. For each question, annotators select the preferred response based on three criteria:~(i) \textbf{Faithfulness:} Does the model answer the question correctly without hallucinating non-existent visual content?~(ii) \textbf{Logical coherence:} Does the reasoning follow a consistent and logical flow?~(iii) \textbf{Tool efficiency:} Does the model arrive at the answer using \zoom efficiently?
On average, \ours-RL is preferred in 59\% of the cases, with a tie in 14\%, indicating the effectiveness of our data generation approach.
Detailed instructions and per-annotator breakdowns are provided in~\cref{appendix:experiment_human_eval}.
Besides human study, we also present a qualitative case study in~\cref{appendix:experiment_eval_traces}. 

\subsection{Ablation Study}
\label{sec:ablation}

\myparagraph{Ablation on question difficulty.}
To investigate the impact of question difficulty on SFT training, we split the 7.1k-sample dataset into an easier half and a harder half based on \qwenvl's \avg.
As shown in~\cref{tab:ablation_qa} (Appendix), the VLM trained on the harder split achieves better performance on \vstar (+2.09) and \visualp-Hard (+4.72), while showing only a slight drop on the easy and medium subsets of \visualp.
These highlight the importance of synthesizing more challenging questions, as described in \cref{sec:qa_synthesis}, to help the model acquire the ability to tackle more complex tasks.

\myparagraph{Ablation on logic chain-driven reasoning trace synthesis.}
To examine the impact of explicitly generating logic chains for each question to guide reasoning trace synthesis, we compare our method to an alternative that uses only fixed in-context examples based on multi-hop and exhaustive search strategies.
As shown in~\cref{tab:ablation_chain} in the Appendix, incorporating logic chains improves overall performance after SFT and yields a substantial gain of 2.87 points after RL.
These results suggest that logic chain-driven synthesis is essential for creating higher-quality SFT data that effectively incentivizes logical reasoning and tool-use capabilities in VLMs.

\myparagraph{Ablation on models used for data synthesis.}
To verify that our data engine's effectiveness stems from the framework design rather than specific large models, we replace the QA generator, the VLM used for QA judging and trace synthesis, and the logic chain generator in~\cref{sec:logt} with much smaller variants: \qwenvlnew~\citep{bai2025qwen3vltechnicalreport}, \qwenvlmid~\citep{bai2025qwen2}, and \gptsmall~\citep{openai2025gptoss}, respectively.
We then re-synthesize 4k SFT samples, train the base model with SFT, and evaluate it on \vstar, \visualp, and \hrbench. As shown in~\cref{tab:ablation_models} in the Appendix, performance remains nearly unchanged compared with using the original models for synthesizing 4k SFT data.

Using these smaller models, generating 7.1k SFT samples requires about 120 GPU hours (30 hours on 4×A100), making it over 25× compute cheaper than the RL training of \minio.

%% file: tables/main_results.tex
\begin{table}[t]
\small
\centering
\caption{\textbf{Performance comparison between our models and baselines.}
We evaluate the \pass accuracy of various models on seven test sets across four benchmarks, and use \mme to denote the out-of-domain benchmark \mmelong.
As checkpoints from Li~\etal~\citep{li2025looklessreasonmore} are not publicly available yet, we report evaluation results directly from the original paper and leave unavailable results blank.
For open-sourced SFT and RL models, the best and second-best results are marked in \textbf{bold} and \underline{underlined}, respectively.
The {\setlength{\fboxsep}{1pt}\colorbox{green!25}{green}} rows refer to our models.
Using only 5.4k–7.1k samples from the SFT dataset synthesized by \ours, the resulting \ours-SFT outperforms existing baselines, including those trained with both SFT and RL.
After applying an off-the-shelf RL recipe without any tool-specific design (see~\cref{sec:experiment_implementation}), \ours-RL surpasses all open-sourced baselines by at least 9.59 points.}

\begin{adjustbox}{max width=\columnwidth}
\begin{tabular}{lccccccccccc}
\toprule 
 & \multicolumn{2}{c}{Data} & \multirow{2}{*}{\vstar} & \multicolumn{3}{c}{\visualp} & \multicolumn{2}{c}{\hrbench} & \multirow{2}{*}{\begin{tabular}[c]{@{}c@{}}MME- \\ RW-Lite\end{tabular}} & \multirow{2}{*}{Avg.} \\
\cmidrule(lr){2-3}\cmidrule(lr){5-7}\cmidrule(lr){8-9}
 & SFT & RL &  & Easy & Medium & Hard & 4K & 8K &  &  \\
\midrule

\rowcolor{gray!20}
\multicolumn{11}{l}{\textit{Models}} \\
GPT-4o         & -   & -    & 66.00 & 47.50 & 15.40 & 11.20 & 59.00 & 55.50 & 46.40 & 43.00 \\
o3             & -   & -    & 95.70 & 69.50    & 47.01    & 32.08    & -    & -    & -    & - \\
\midrule

\rowcolor{gray!20}
\multicolumn{11}{l}{\textit{Open-sourced Models w/o Tools}} \\
\qwenvl  & -   & -    & 79.58 & 41.81 & 23.13 & 19.81 & 72.38 & 66.25 & 48.46 & 50.20 \\
\qwenvlxl & -   & -    & 79.06 & 39.72 & 32.09 & 28.30 & 76.25 & 73.00 & 50.13 & 54.08 \\
\midrule

\rowcolor{gray!20}
\multicolumn{11}{l}{\textit{Open-sourced SFT Models with Tools}} \\
\pixel-SFT & 4k & 0     & 74.87 & 49.65 & 18.66 & 20.75 & 64.75 & 59.88 & 47.06 & 47.95 \\
\cof-SFT & 5.4k & 0     & 83.77 & 53.90 & 27.61 & 34.91 & 71.50 & 65.25 & \underline{48.05} & 55.00 \\
\minio-SFT        & 7.3k & 0     & 64.40 & 32.62 & 28.73 & 14.15 & 53.75 & 51.75 & 38.56 & 40.57 \\

\rowcolor{green!5}
\ours-SFT           & 4k & 0     & 85.34 & \textbf{61.70} & 34.70 & 36.79 & 71.50 & 67.25 & 43.93 & 57.32 \\
\rowcolor{green!5}
\ours-SFT           & 5.4k & 0     & \underline{85.86} & \underline{60.99} & \underline{38.06} & \underline{37.74} & \textbf{72.25} & \textbf{70.62} & \textbf{48.15} & \underline{59.10} \\
\rowcolor{green!5}
\ours-SFT           & 7.1k & 0     & \textbf{87.43} & \underline{60.99} & \textbf{41.04} & \textbf{39.62} & \underline{72.00} & \underline{70.12} & 45.96 & \textbf{59.59} \\
\midrule

\rowcolor{gray!20}
\multicolumn{11}{l}{\textit{Open-sourced RL Models with Tools}} \\
\pixel-RL & 4k & 7k  & 81.68 & 58.87 & 24.25 & 26.42 & 73.00 & 67.38 & 52.42 & 54.86 \\
\cof-RL & 5.4k & 8.9k    & 83.25 & 55.32    & 28.36    & \underline{36.79}    & 72.50    & 65.88    & 48.83    & 55.85 \\
\deep-RL       & 0    & 47k & 83.25 & \underline{62.41} & 27.99 & 30.19 & \underline{74.62} & \underline{69.50} & \textbf{53.47} & 57.35 \\
Li~\etal~\citep{li2025looklessreasonmore}      & 4k & 7k  & 85.86 & -    & -    & -    & 73.40 & 66.60 & -    & - \\
\minio-RL        & 7.3k & 9.7k  & \underline{89.53} & 51.77 & \underline{45.15} & \underline{36.79} & 71.25 & 66.62 & 42.73 & \underline{57.69} \\
\rowcolor{green!5}
\ours-RL           & 7.1k & 6.4k  & \textbf{92.15} & \textbf{70.92} & \textbf{52.24} & \textbf{51.89} & \textbf{76.12} & \textbf{74.62} & \underline{53.00} & \textbf{67.28} \\
\rowcolor{green!5}
$\Delta$ vs. SOTA          & - & -     & \Inc{2.62} & \Inc{8.51} & \Inc{7.09} & \Inc{15.10} & \Inc{1.50} & \Inc{5.12} & \Dec{0.47} & \Inc{9.59} \\
\rowcolor{green!5}
$\Delta$ vs. Base Model & - & -     & \Inc{12.57} & \Inc{29.11} & \Inc{29.11} & \Inc{32.08} & \Inc{3.74} & \Inc{8.37} & \Inc{4.54} & \Inc{17.08} \\
\bottomrule
\end{tabular}
\label{tab:main_results}
\end{adjustbox}
\end{table}

%% file: figures/rl_and_analysis.tex
\begin{figure}[t]
    \centering
    \begin{minipage}{0.51\linewidth}
        \centering
        \includegraphics[width=0.9\linewidth]{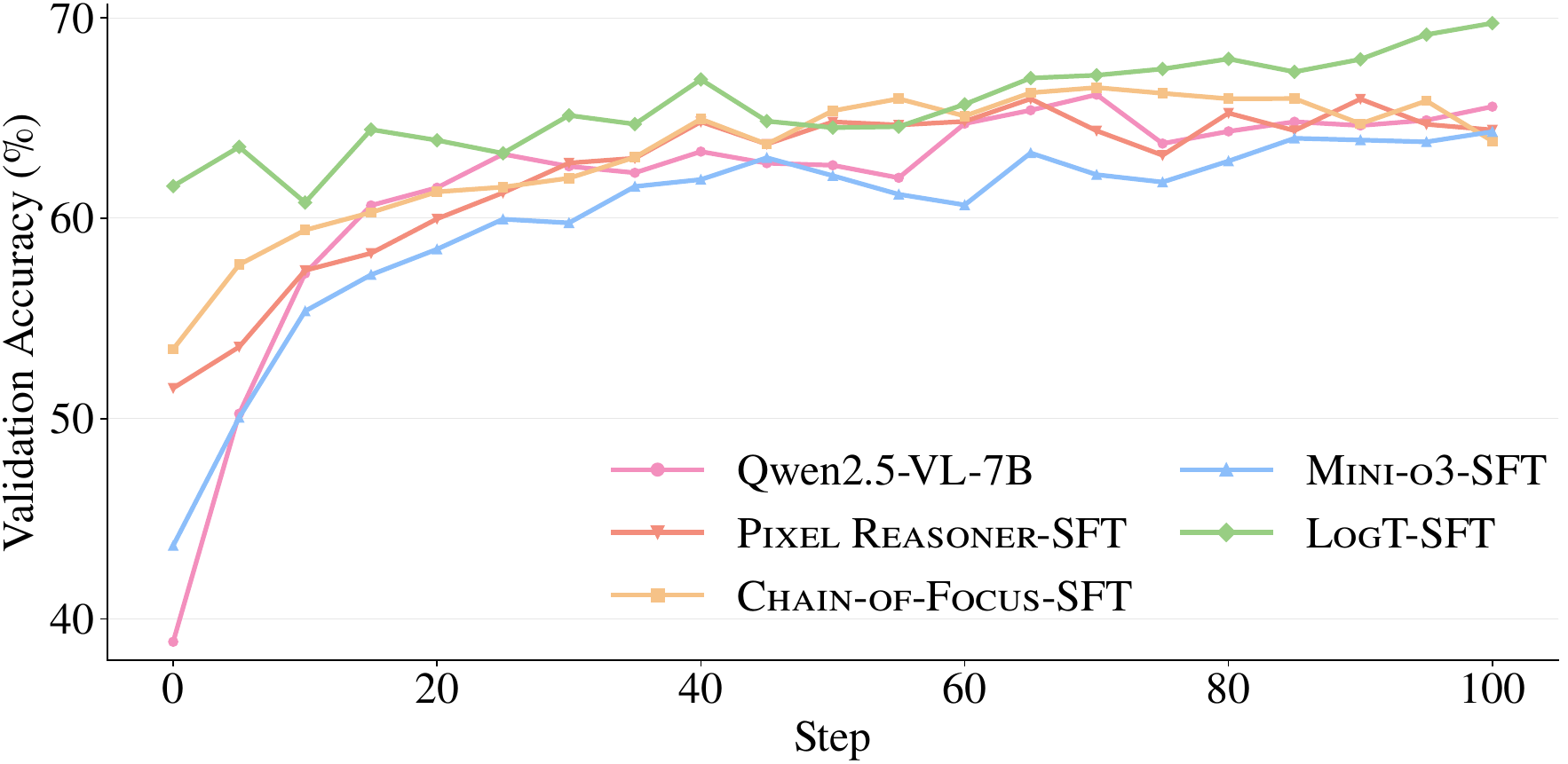}
    \caption{\textbf{Training dynamics after applying our simple RL procedure to the baselines and \ours-SFT.}
    During RL, our model's validation accuracy steadily improves, whereas other SFT baselines perform on par with, or worse than zero-RL. \ours greatly improves the RL performance ceiling.}
    \label{fig:rl_dynamics}
    \end{minipage}
    \hfill
    \begin{minipage}{0.46\linewidth}
        \centering
        \includegraphics[width=\linewidth]
    {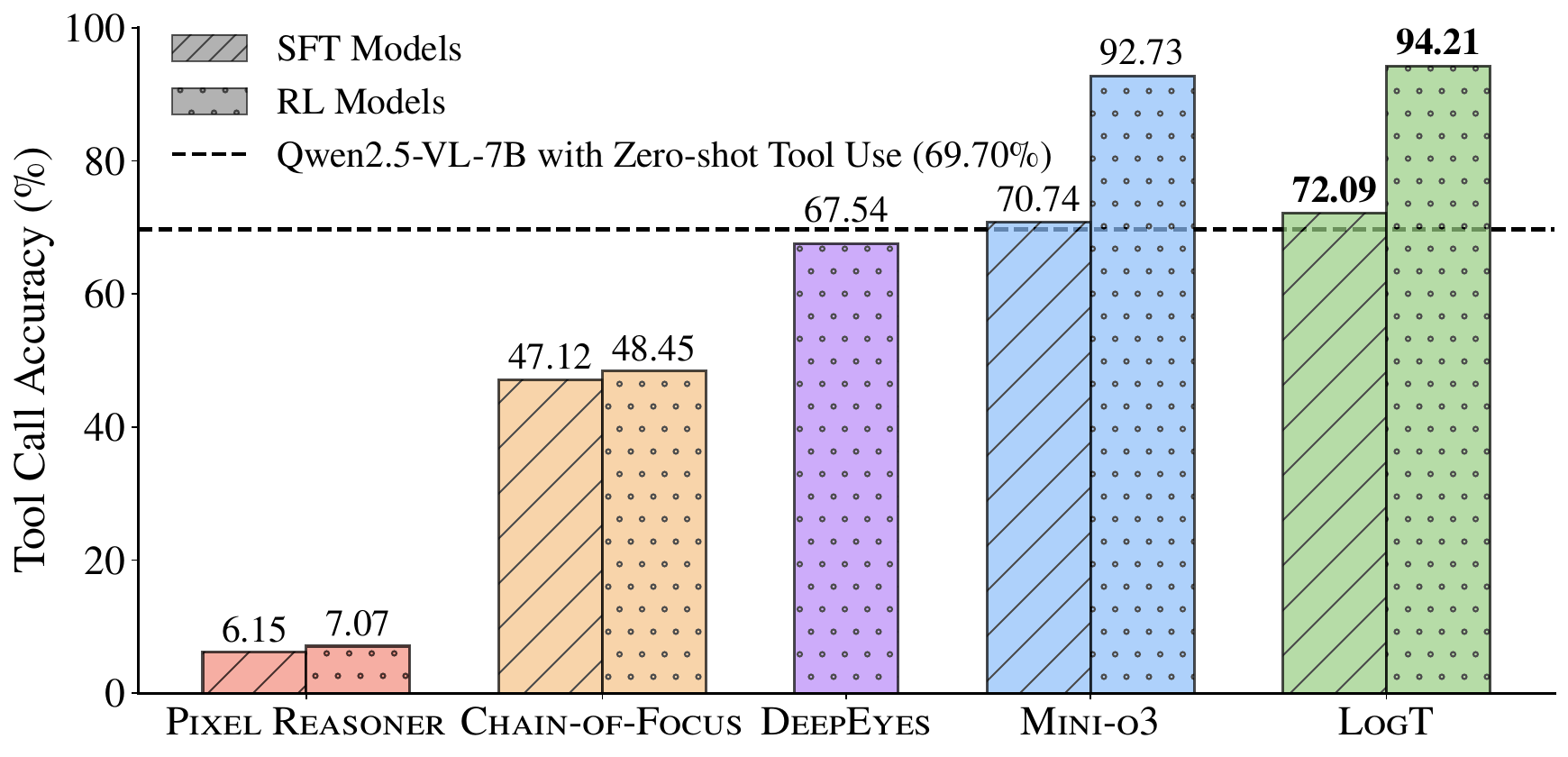}
    \caption{\textbf{Tool call accuracy.} We report tool call accuracy on \vstar for the base model and all post-trained models. The base model is prompted to use \zoom in a zero-shot setting. Our \ours-SFT and \ours-RL models achieve the highest \zoom accuracy.}
    \label{fig:tool_accuracy}
    \end{minipage}
\end{figure}

%% file: figures/tool_efficiency_figure.tex
\begin{figure}[t]
    \centering
    \includegraphics[width=0.6\linewidth]{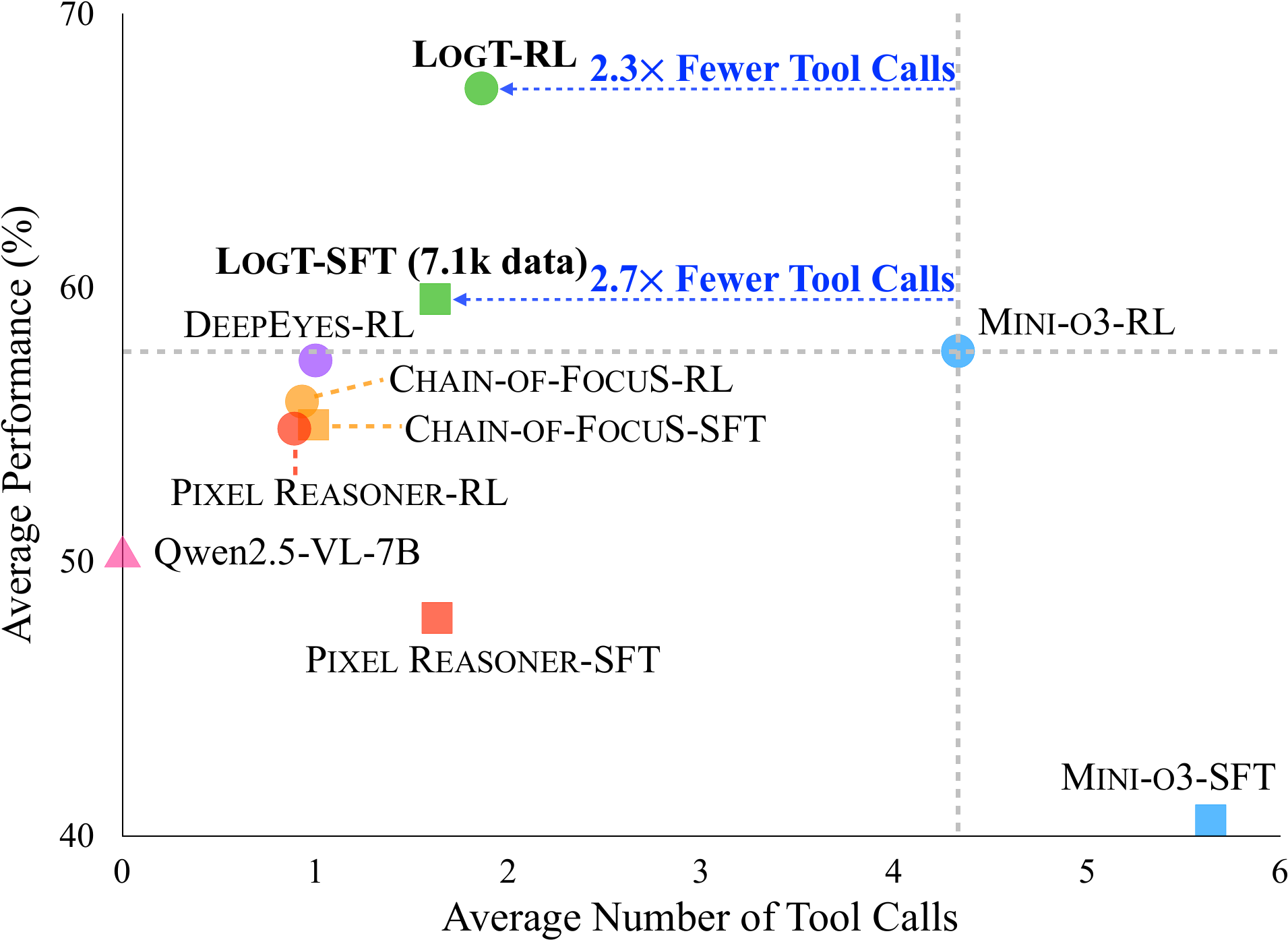}
    \caption{\textbf{Tool call efficiency.} We report the average number of tool calls per question across the seven test sets. Despite achieving better performance compared to prior SOTA, \ours-SFT model uses 2.7$\times$ fewer tool calls, and \ours-RL uses 2.3$\times$ fewer.}
    \label{fig:tool_efficiency}
\end{figure}

%% file: sec/5_discussion.tex
\section{Discussion}
\label{sec:discussion}

\subsection{Connection to Cognitive Science}
The reasoning strategies in~\cref{sec:trace_synthesis} have close parallels in findings from cognitive science:~(i)
\textbf{Direct answering.}
Feature-integration theory~\citep{treisman1980feature} proposes that humans can detect simple visual features in parallel with no attention limits, enabling an immediate impression of recognizable objects.
This is analogous to a VLM directly answering a visual question without invoking tools to inspect finer details.~(ii)
\textbf{Multi-hop.}
Research on natural vision~\citep{rao2002eye} suggests that visual information is acquired actively, in a task-dependent and coarse-to-fine manner: observers first attend to a promising contextual region and then inspect task-relevant details.
This process closely parallels the multi-hop reasoning strategy used in our work.~(iii)
\textbf{Exhaustive search.}
Guided-search models~\citep{wolfe2017five} propose that visual search is efficient when target features, scene context, or prior knowledge prioritize a small subset of locations.
When these signals are weak, attention must cover more candidates, resulting in a more systematic search. This corresponds to the exhaustive-search strategy used in our work.

\subsection{Relationship to Knowledge Distillation}
The model ablation (\cref{tab:ablation_models}) suggests that \ours does not rely on the original large synthesizing models.
Moreover, on \vstar, the 7B model trained on 4k reasoning traces synthesized using \qwenvlmid outperforms \qwenvlmid itself by 4.18 points.
Together, these results suggest that synthesizing-model strength alone does not explain the effectiveness of our data engine, although knowledge distillation~\citep{fu2025tshirt, hamman2025fewshot} may still contribute to the gains.

\subsection{Limitations}
\label{sec:limitations}

\myparagraph{The centrality of \zoom.}
In this work, we focus primarily on \zoom for two primary reasons.~(i)
\textbf{Practical importance.}
The zoom-in tool is widely used in visual reasoning tasks.
For example, OpenAI's o3 model prominently features this tool in its blog~\citep{openai2025thinking}.
Moreover, all thinking-with-image baselines in~\cref{tab:main_results} rely on zoom-in as their sole image interaction mechanism.~(ii)
\textbf{Technical challenge.}
Despite its prevalence, enabling VLMs to accurately and efficiently decide when and how to invoke the zoom-in tool remains challenging (see analysis in~\cref{sec:analysis}).
OpenAI's blog~\citep{openai2025thinking} similarly identifies ``redundant or unnecessary tool calls and image manipulation steps'' as a key limitation of o3.

While centered on \zoom, our approach demonstrates significant transferability.
\ours improves performance on out-of-domain tasks (\eg, \mme in \cref{tab:main_results}) and generalizes to unseen tools (\cref{tab:rotation} in~\cref{appendix:experiment_rotation}).

More broadly, \ours decouples logic chain synthesis from reasoning trace generation, making it a general and extensible framework.
This design allows it to be applied to a wide range of structured reasoning tasks, and extending \ours to broader domains is a promising direction for future work.

\myparagraph{Reliance on VLM-as-a-Judge.}
To ensure scalability and avoid costly human annotation, we build a fully automated data engine that uses VLM-as-a-Judge.
Consequently, the synthesized SFT data may contain a small amount of noise, as VLM judges can make mistakes.
We view this trade-off as acceptable for two primary reasons.~(i)
\textbf{Annotation noise is unavoidable.}
Errors are not unique to automated judging; they are often inevitable even in expensive human annotations~\citep{vendrow2025largelanguagemodelbenchmarks, openai2026why}.~(ii)
\textbf{Empirical results and manual verification suggest limited noise.}
Our method achieves the strongest empirical performance (\cref{tab:main_results}) and the highest tool-call accuracy (\cref{fig:tool_accuracy}), suggesting that any remaining noise is limited and much lower than in prior work.
We further randomly sample 100 examples and manually evaluate errors at each stage of the data engine.
As shown in~\cref{tab:error_rate}, the error rate at each stage is low despite the fully automated pipeline.

%% file: sec/6_conclusion.tex
\section{Conclusion}
We propose \ours, a novel data engine for synthesizing high-quality SFT data that empowers VLMs to logically think with images and use \zoom both effectively and efficiently.
The core is to generate challenging QAs and high-quality tool-use reasoning traces, guided by logic chains that emulate human-like visual reasoning strategies.
VLMs fine-tuned on our synthetic data outperform existing baselines trained with both SFT and RL.
Moreover, applying off-the-shelf RL on top of \ours-SFT, without any tool-specific design, further substantially improves performance.

%% file: sec/ack.tex
\section*{Acknowledgment}
We would like to thank Hugo Lauren\c{c}on, Jianwei Yang, Kaiwen Zha, and Xingyi Zhou for their helpful discussions during this project.

%% file: sec/appendix.tex
\clearpage
\appendix
\begingroup
\raggedbottom
\setlength{\intextsep}{8pt plus 2pt minus 2pt}
\widowpenalty=10000
\clubpenalty=10000

\renewcommand{\topfraction}{0.90}
\renewcommand{\textfraction}{0.10}
\renewcommand{\floatpagefraction}{0.65}
\setcounter{topnumber}{6}
\setcounter{totalnumber}{6}
\setlength{\floatsep}{10pt plus 2pt minus 2pt}
\setlength{\textfloatsep}{12pt plus 2pt minus 2pt}
\makeatletter
\setlength{\@fptop}{0pt}
\setlength{\@fpsep}{12pt plus 1fil}
\setlength{\@fpbot}{0pt plus 2fil}
\makeatother

\section*{Appendix}

\section*{Contents}
\begin{itemize}
  \item \textbf{Section}
  \hyperref[appendix:approach]{\textbf{A}}: Additional Details on Our Data Engine: \ours
  \begin{itemize}
      \item \textbf{Section} \hyperref[appendix:approach_prompt]{\textbf{A.1}}: System Prompts for QA Generator and Reasoning Trace Generator
      \item \textbf{Section} \hyperref[appendix:approach_exhaustive]{\textbf{A.2}}: An Example of Exhaustive Search Reasoning Trace
      \item \textbf{Section} \hyperref[appendix:approach_tool_call_distribution]{\textbf{A.3}}: Tool Use Statistics in the \ours SFT Dataset
      \item \textbf{Section} \hyperref[appendix:approach_trace_comparision]{\textbf{A.4}}: Qualitative Case Study of Synthesized Tool-use Traces
      \item \textbf{Section} \hyperref[appendix:approach_costs]{\textbf{A.5}}: Computational Cost Analysis
  \end{itemize}

  \item \textbf{Section} \hyperref[appendix:experiment]{\textbf{B}}: Additional Experiment Setup and Results
  \begin{itemize}
      \item \textbf{Section} \hyperref[appendix:experiment_eval_details]{\textbf{B.1}}: Additional Evaluation Details
      \item \textbf{Section} \hyperref[appendix:experiment_rl_details]{\textbf{B.2}}: Additional RL Implementation Details
      \item \textbf{Section} \hyperref[appendix:experiment_baselines]{\textbf{B.3}}: Performance Comparison Between \ours and Additional Baselines (Including \sketch, \pyvision, \internvl, \llavaov, \seal, and \dyfo)
      \item \textbf{Section} \hyperref[appendix:experiment_internvl]{\textbf{B.4}}: SFT Results on an Additional Base Model: \internvl
      \item \textbf{Section} \hyperref[appendix:experiment_human_eval]{\textbf{B.5}}: Additional Human Evaluation Details
      \item \textbf{Section} \hyperref[appendix:experiment_rotation]{\textbf{B.6}}: Zero-Shot Generalization to Unseen Tools
      \item \textbf{Section} \hyperref[appendix:experiment_eval_traces]{\textbf{B.7}}: Qualitative Case Study of Model Responses Generated By Baselines and \ours-RL
  \end{itemize}
\end{itemize}

\section{Additional Details on Our Data Engine: \ours}
\label{appendix:approach}

\subsection{System Prompts for QA Generator and Reasoning Trace Generator}
\label{appendix:approach_prompt}

We detailed the QA synthesis in~\cref{sec:qa_synthesis} and the reasoning trace synthesis in~\cref{sec:trace_synthesis}.
Here, we show part of the system prompt for the QA generator in~\cref{fig:qagen_prompt} and part of the system prompt for the reasoning trace generator in~\cref{fig:tracegen_prompt}.

\input{figures/qagen_prompt}
\input{figures/tracegen_prompt}

\subsection{An Example of Exhaustive Search Reasoning Trace}
\label{appendix:approach_exhaustive}

We provide an additional example of an exhaustive-search logic chain and its corresponding tool-use reasoning trace in~\cref{fig:exhaustive_search}.
In this example, the target object for the given question, \ie, the pink sign held by the person, is particularly difficult to locate.
Therefore, the logic chain selector (see~\cref{sec:trace_synthesis}) chooses the exhaustive search logic chain.
This logic chain instructs the reasoning trace generator to first zoom in on the crowd and then systematically search the scene from left to right. 
Although the logic chain specifies examining all four regions of the crowd (left, center-left, center-right, and right) because the logic chain generator does not know exactly where the target object is located, the reasoning trace generator successfully identifies the target object after zooming into the second region (center-left crowd).
Consequently, it terminates the search early.
This is because our prompt requires the reasoning trace generator to stop searching once the target object is found.
After identifying the target, it calls \zoom on the object and then proceeds to generate the final answer.

\input{figures/exhaustive_search_figure}

\begin{figure}[!t]
\centering
\begin{minipage}[t]{0.48\linewidth}
\input{figures/sft_tool_distribution_figure}
\end{minipage}\hfill
\begin{minipage}[t]{0.48\linewidth}
\input{figures/easy_questions_figure}
\end{minipage}
\end{figure}

\subsection{Tool Use Statistics in the \ours SFT Dataset}
\label{appendix:approach_tool_call_distribution}

We present the normalized distribution of \zoom call counts in~\cref{fig:sft_tool_distribution}.
The number of tool calls ranges from 0 to 6.
Specifically, 94\% of the questions in the \ours SFT dataset are solved using 3 or fewer \zoom calls.

\subsection{Qualitative Case Study of Synthesized Tool-use Traces}
\label{appendix:approach_trace_comparision}

We present a qualitative comparison of reasoning trace synthesis strategies used by \pixel~\citep{su2025pixel}, \cof~\citep{zhang2025chain}, \minio~\citep{lai2025mini}, and \ours, as shown in~\cref{fig:sft_trace_comparison}.

\pixel synthesizes multi-turn reasoning traces by deliberately inserting \zoom calls that focus on overly broad or incorrect regions.
In~\cref{fig:pixel_vs_logt}, the first two \zoom calls target completely irrelevant areas.
In contrast, \ours generates a logic chain that guides the reasoning trace generator to first invoke \zoom on the building, then refine the focus onto the target text.

In~\cref{fig:cof_vs_logt}, the \cof SFT trace redundantly zooms in on the shoes for six consecutive steps, despite correctly identifying their location in the first step.
In contrast, \ours first generates a logic chain that instructs the reasoning trace generator to initially zoom in on the person, then gradually focus on the shoes, and finally answer the question using only two \zoom calls.

A similar issue occurs with the \minio SFT trace in~\cref{fig:minio3_vs_logt}: although it initially identifies the correct target, it repeatedly invokes \zoom and ultimately selects an unrelated region, resulting in a total of four \zoom calls.
By comparison, \ours guides the reasoning trace generator to directly locate the wheels, with the reasoning trace requiring only a single \zoom call before answering.

\input{figures/sft_trace_comparison}

\subsection{Computational Cost Analysis}
\label{appendix:approach_costs}

Synthesizing 7.1k samples with \ours requires approximately 960 GPU hours (5 days on 8 A100 GPUs), making it more than 3$\times$ compute cheaper than RL training for the Mini-o3 baseline. 

As shown in~\cref{sec:ablation}, synthesis resource use can be substantially reduced by using smaller model variants without compromising performance. With smaller models, the total resource use drops to approximately 120 GPU hours, making it more than 25$\times$ compute cheaper than RL training for Mini-o3. 

\clearpage
\input{tables/other_baselines}

\section{Additional Experiment Setups and Results}
\label{appendix:experiment}

\subsection{Additional Evaluation Details}
\label{appendix:experiment_eval_details}

During evaluation, we adopt a maximum image resolution budget of 8192$\times$28$\times$28 pixels for models in the \qwenvl family, where each 28$\times$28 patch represents a visual token.
Images exceeding this budget are downsampled.
For comparison, \minio uses a smaller default budget of approximately 2551$\times$28$\times$28 pixels.
We observe that \minio consistently performs better under our higher budget. 
Therefore, we adopt our budget for all \qwenvl-based evaluations.

\subsection{Additional RL Implementation Details}
\label{appendix:experiment_rl_details}
We provide additional details on our RL implementation that are not covered in~\cref{sec:experiment_implementation}.
During RL, to reduce GPU idle time and improve training efficiency, we cap the number of \zoom calls at 6.
Our RL training data comes from \minio,
and we filter out trivial cases by removing any question where \qwenvl achieves an \avg of 100\%.
To monitor training dynamics during RL, we use a validation set composed of samples from \vstar, \visualp, and \hrbench.

\subsection{Performance Comparison Between \ours and Additional Baselines}
\label{appendix:experiment_baselines}

To further demonstrate the effectiveness of \ours, we evaluate our approach against a broader set of baselines.
Specifically, we include two more open-source models of comparable scale (7B–8B): \internvl~\citep{wang2025internvl3_5} and \llavaov~\citep{li2024llavaonevision}.
We also consider modular-agent frameworks that enable proprietary models to use external tools, such as \sketch~\citep{hu2024visual} and \pyvision~\citep{zhao2025pyvision}.
In addition, we evaluate two visual-search-based methods that rely on external visual experts: \seal~\citep{wu2023vstar} and \dyfo~\citep{li2025dyfo}.
The performance of these baselines, along with \ours-SFT and \ours-RL, is shown in~\cref{tab:other_baselines}.
Results indicate that both \ours-SFT and \ours-RL significantly outperform the additional baselines, consistent with the findings in~\cref{tab:main_results}.

\input{tables/internvl}

\input{figures/humaneval_prompt}

\input{figures/humaneval_breakdown_figure}

\subsection{SFT Results on an Additional Base Model: \internvl}
\label{appendix:experiment_internvl}

To evaluate whether our data engine generalizes to other base models, we fine-tune \internvl using 7.1k SFT samples generated by \ours. As shown in~\cref{tab:internvl}, the fine-tuned model substantially outperforms the base model, achieving a 14-point overall improvement.

\subsection{Additional Human Evaluation Details}
\label{appendix:experiment_human_eval}
We present the instructions given to human annotators in~\cref{fig:humaneval_prompt}.
For each question, the evaluation interface randomly assigns \minio-RL and \ours-RL to either Model $A$ or Model $B$.
In addition to the overall human evaluation results reported in~\cref{sec:analysis}, the per-annotator breakdown in~\cref{fig:huameval_breakdown} shows that all annotators prefer responses generated by \ours-RL over those from the prior SOTA, \minio-RL.
This underscores that our data generation framework enables VLMs to reason more effectively using the \zoom tool.

\input{tables/ablation_qa}
\input{tables/ablation_chain}
\input{tables/ablation_models}
\input{tables/error_rate}

\input{tables/zero_shot_rotation}

\subsection{Zero-Shot Generalization to Unseen Tools}
\label{appendix:experiment_rotation}
To evaluate whether our method can generalize to tools unseen during training, we construct a new benchmark, \vstar-Rotated, by rotating images in \vstar by angles uniformly sampled from $30^{\circ}$ to $330^{\circ}$.
We evaluate the base model and \ours-RL on this benchmark. At inference time, \ours-RL is given access to a rotation tool that was not seen during training. As shown in~\cref{tab:rotation}, \ours-RL learns to use this tool in a zero-shot setting despite receiving no rotation-related supervision, outperforming the base model by 21.99 points.
These results suggest that \ours can generalize to novel tasks and previously unseen tools.

\input{figures/eval_trace_comparison_figure}

\subsection{Qualitative Case Study of Model Responses Generated By Baselines and \ours-RL}
\label{appendix:experiment_eval_traces}

To illustrate how SFT data synthesized by \ours helps VLMs reason logically and use \zoom accurately, we present a qualitative case study comparing model responses to a question from the \vstar benchmark.
These comparisons are shown in~\cref{fig:eval_trace_comparison}.
We refer to \qwenvl-RL as the model obtained by directly applying our RL training procedure from~\cref{sec:experiment_implementation} to the base model \qwenvl.
For all other baselines, we use the officially released checkpoints.

\begin{itemize}[leftmargin=2em, topsep=0pt, itemsep=0pt, parsep=0pt]
\item \textbf{\qwenvl-RL} answers the question correctly but hallucinates non-existent white text on the tissue box.
\item \textbf{\pixel-RL} also produces the correct answer; however, its \zoom call targets an irrelevant region, and its final answer contradicts its own statement that ``There is no tissue box.''
\item \textbf{\deep-RL} focuses on white sheets of paper and incorrectly concludes that the tissue box is white.
\item \textbf{\cof-RL} zooms in on an irrelevant region and hallucinates, though it ultimately answers correctly.
\item \textbf{\minio-RL} selects regions that include the tissue box, but still arrives at an incorrect answer.
\item \textbf{\ours-RL} first identifies the likely location of the tissue box (the truck's front cabin), then zooms in on the tissue box itself, ultimately producing the correct answer without any hallucinations.
\end{itemize}
Among all the models, \ours-RL is the only one that reaches the correct answer and shows no hallucinations.
This example further demonstrates that our data engine produces high-quality SFT data that effectively trains VLMs to reason about images in a logical and faithful manner.

\clearpage
\endgroup

%% file: figures/qagen_prompt.tex
\begin{figure}[!tb]
\centering
\begin{qagenbox}{System Prompt for QA Generator}
\begin{lstlisting}
You are a Vision-Language assistant tasked with generating one high-quality question and corresponding answer for each input image. The purpose is to synthesize QA datasets based on visual content from specific regions in the image.

## Inputs (per task)
1. The input image, with key regions marked by red boxes and larger surrounding areas marked by blue boxes.
2. Annotations, including:
   - Input image size.
   - One or more bounding boxes (indicating key regions).
   - Larger cropped images that **include** the red bounding boxes.
   - Cropped images that **correspond to** the red bounding boxes.

## Your task
1. Analyze the full image, the marked regions, and the cropped images.
2. Determine whether the input image and the provided region(s) provide enough information to generate a high-quality question.
3. If yes, generate one question and its corresponding brief answer.
4. If not, return {(*@``@*)valid(*@''@*): false}.

## Rules
1. **Location**
The question must be about the object(s) **within** the provided **red** bounding box(es).
   - **Be very careful**: the target object(s) tied to the answer must lie **within the red bounding box**, not merely nearby and not just within the larger blue box(es).
2. **Type**
Each question must fit one of the following types:
   - (*@``@*)perception_single(*@''@*): attribute of a single object (e.g., color, category, material).
   - (*@``@*)perception_multi(*@''@*): relationship between multiple objects (e.g., larger/smaller, left/right).
   - (*@``@*)counting(*@''@*): count of objects in the **entire image**, but only if those objects appear in at least one bounding box.
   - (*@``@*)ocr(*@''@*): read visible text in the region(s) (e.g., words, numbers, labels).
   - (*@``@*)reasoning(*@''@*): interpret visuals that require math or chart understanding.
(The remaining 5 longer rules are omitted for brevity.)

## Example questions by type
(Example questions are omitted for brevity.)
    \end{lstlisting}
    \end{qagenbox}
    \caption{\textbf{System prompt for the QA generator.} The remaining rules and in-context examples are omitted for brevity.}
    \label{fig:qagen_prompt}
\end{figure}

%% file: figures/tracegen_prompt.tex
\begin{figure}[!tb]
\centering
\begin{tracegenbox}{System Prompt for Reasoning Trace Generator}
\begin{lstlisting}
You are a **visual question answering assistant** equipped with a tool called (*@``@*)zoom_in(*@''@*), which crops a specified region from an image.

## Input (per task)
1. The input image (image_id: 0).
2. Annotations, including:
   - Ground-truth region(s) shown as red boxes in the input image.
   - Input image size.
   - Cropped image(s) corresponding to the ground-truth region(s).
3. Question.
4. Logic chain describing the sequential regions or relations to inspect.

## Your task
1. Your task is to produce a **complete, step-by-step reasoning trace** that visually derives the answer to a question using only what is visible in the provided image.
2. Your reasoning trace must **follow the given logic chain**, though you must never mention the chain itself in your reasoning text.

## Chain structure
1. Each chain is a **numbered sequence of steps**.
2. Each step follows the format <short noun phrase> (k):
   - The **short noun phrase** names the region, object, or relation to inspect.
   - The optional **(k)** is a numeric zoom tag indicating which image_id to zoom into.

## Rules
You must strictly follow all rules.
1. Simulate visual reasoning from scratch and act as if you only have access to:
   - the **original image** (without ground-truth region(s)),
   - and the **question**.
Do **not** mention annotations, ground-truth region(s), or the chain explicitly. Simulate a realistic, step-by-step visual discovery process.
2.1. Follow the logic chain exactly (multi-hop reasoning chain only).
   - Strictly follow every step in the chain **in order**, no skipping, merging, or reordering.
2.2. Be exhaustive but structured. Search systematically, not randomly (exhaustive-search chain only).
   - Four broad patches max when scanning a large area; examine them in a deterministic order.
   - Fully examine each candidate region before moving on.
   - Stop the search only when you confirm the target visually.
(The remaining 8 longer rules are omitted for brevity.)

## Examples
(Example reasoning traces are omitted for brevity.)
\end{lstlisting}
\end{tracegenbox}
\caption{\textbf{System prompt for the reasoning trace generator.}
Each logic-chain entry of the form $\langle\text{short noun phrase}\rangle(k)$ specifies a target object or region and its source image $I_k$.
Here, $I_0$ is the original image, and $I_k$ for $k \geq 1$ is the crop returned by the $k$-th \zoom call.
The reasoning trace generator locates the specified target in $I_k$, predicts its bounding box, and invokes \zoom on that image using the predicted box.
For example, ``Backpack (2)'' corresponds to ``Backpack ($I_2$)'' in~\cref{fig:logt_overview} and instructs the generator to zoom in on the backpack within the person crop returned by the second tool call.
The remaining rules and in-context examples are omitted for brevity.}
\label{fig:tracegen_prompt}
\end{figure}

%% file: figures/exhaustive_search_figure.tex
\begin{figure}[!tb]
    \centering
    \includegraphics[width=\linewidth]{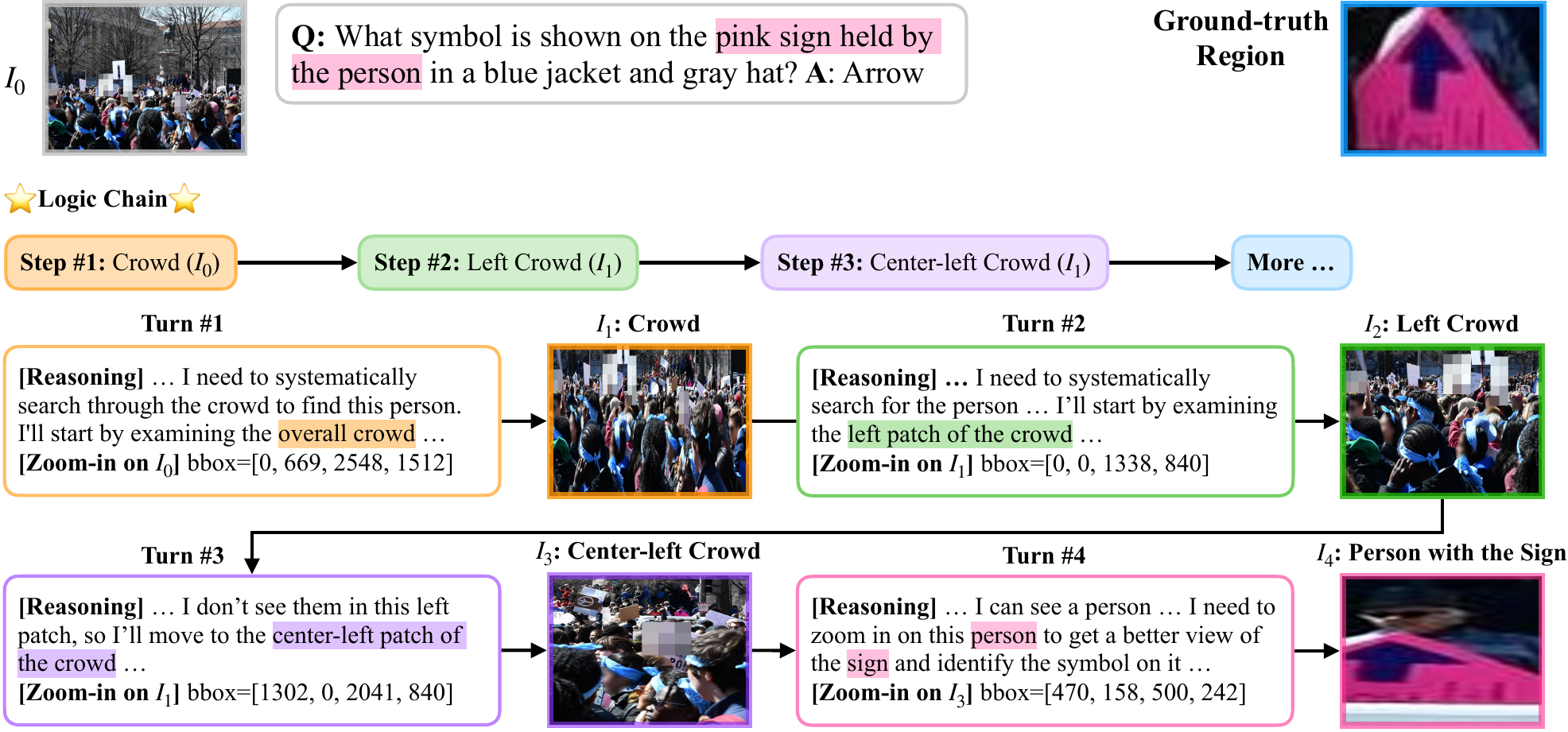}
    \caption{\textbf{An example of an exhaustive search logic chain and its corresponding tool-use reasoning trace.} To locate the pink sign held by a person in the crowd, the logic chain begins by instructing the trace generator (see~\cref{sec:trace_synthesis}) to identify the crowd, then systematically search it from left to right. Since the trace generator successfully detects the target sign and the person in the crop returned by the third \zoom call, it stops scanning the rest of the crowd and instead zooms in on the identified region to extract the final answer.}
    \label{fig:exhaustive_search}
\end{figure}

%% file: figures/sft_tool_distribution_figure.tex
\begin{figure}[H]
    \centering
    \includegraphics[width=\linewidth]{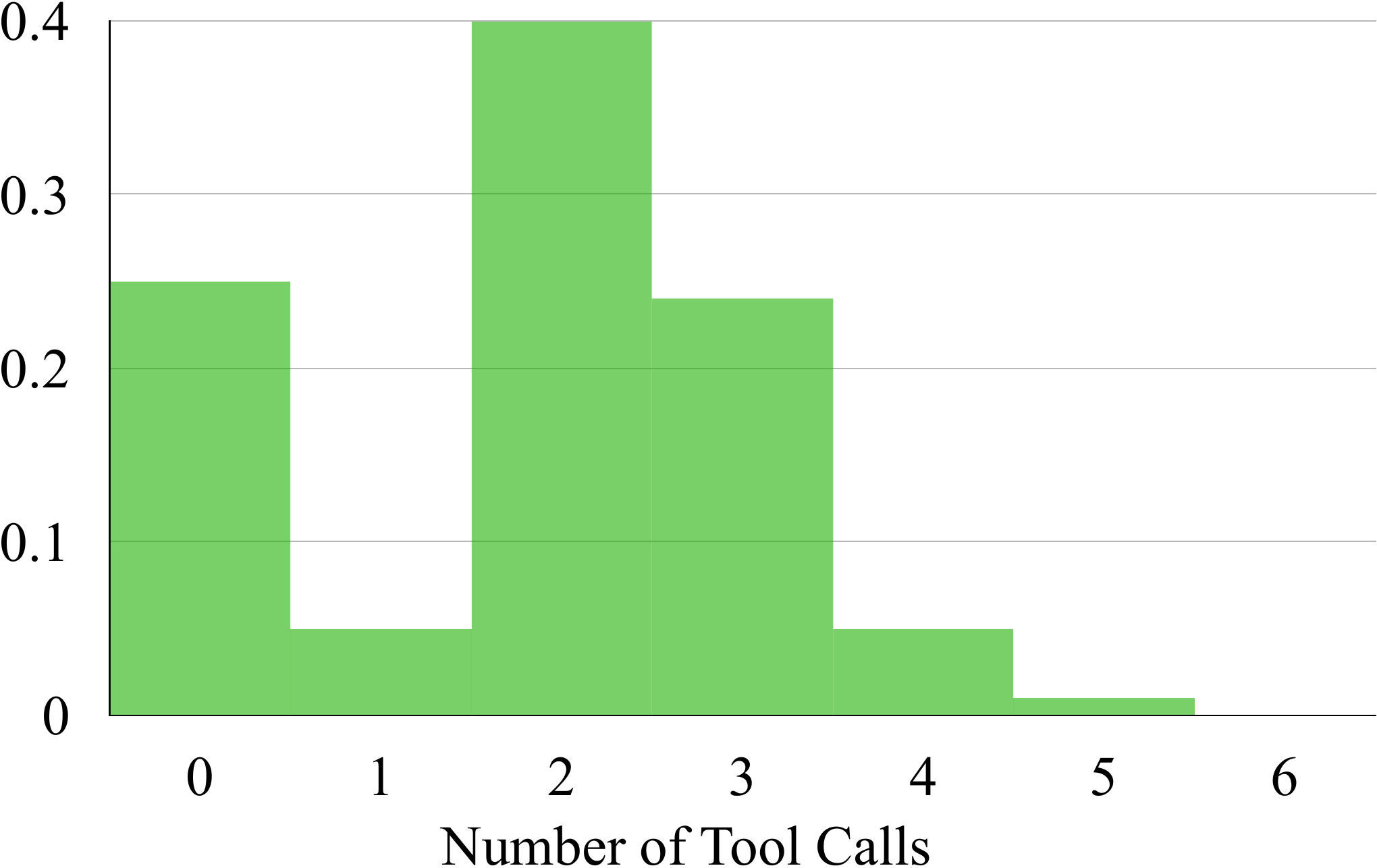}
    \caption{
    \textbf{Normalized distribution of \zoom call counts in the SFT dataset generated by \ours.}
    The number of tool calls per trace in the dataset ranges from 0 (direct answering without tool use) to 6.
    }
    \label{fig:sft_tool_distribution}
\end{figure}

%% file: figures/easy_questions_figure.tex
\begin{figure}[H]
    \centering
    \includegraphics[width=\linewidth]
    {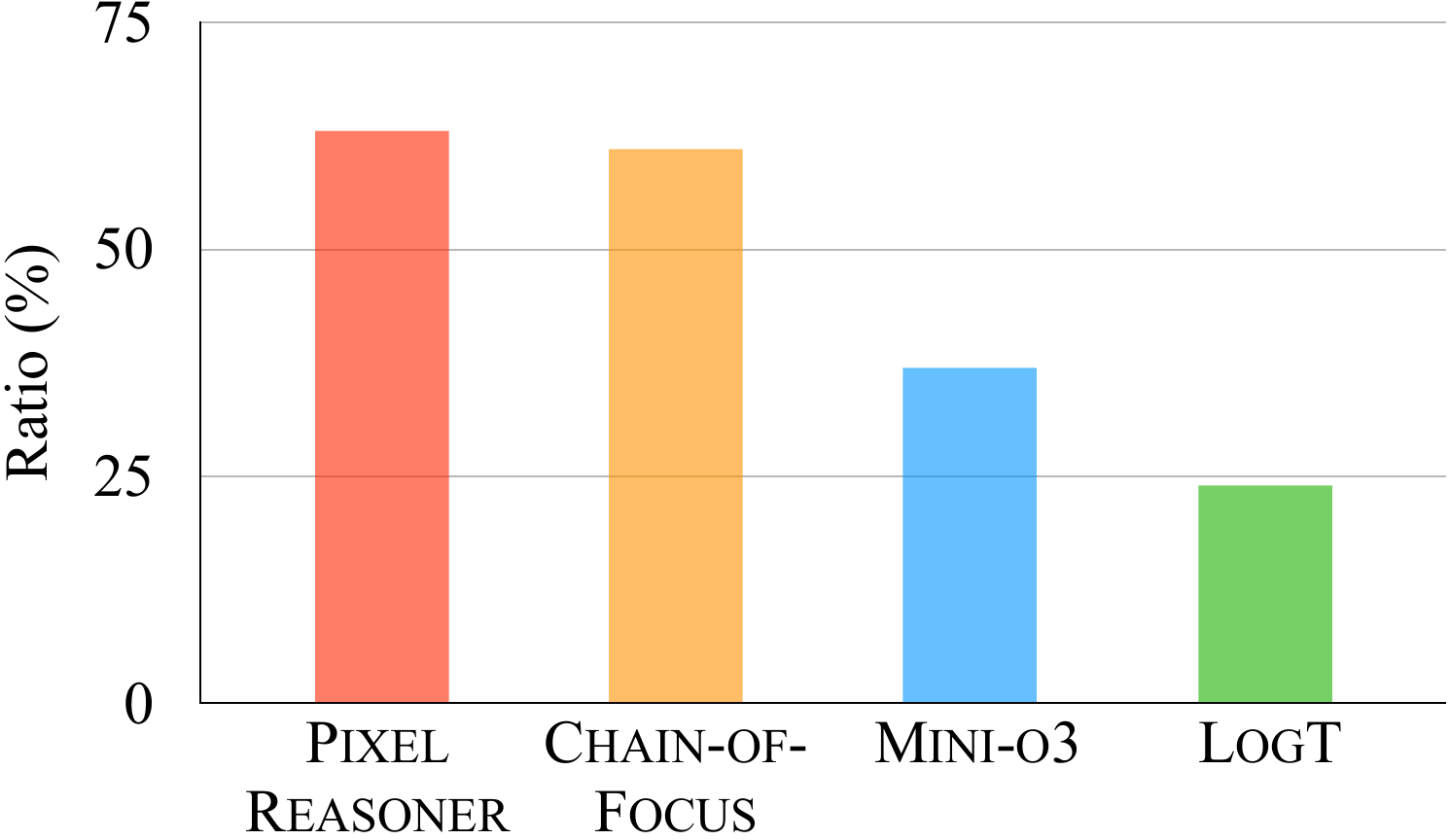}
    \caption{\textbf{Ratio of easy questions.} We compute the ratio of questions in existing and our SFT datasets for which \qwenvl achieves an \avg$=1$ without needing tools (easy questions). Our dataset synthesized by \ours contains only 24\%, significantly lower than existing datasets. 
    }
    \label{fig:easy_questions}
\end{figure}

%% file: figures/sft_trace_comparison.tex
\begin{figure*}[!tb]
    \centering
    \begin{subfigure}[t]{0.75\linewidth}
        \centering
        \includegraphics[width=\linewidth]{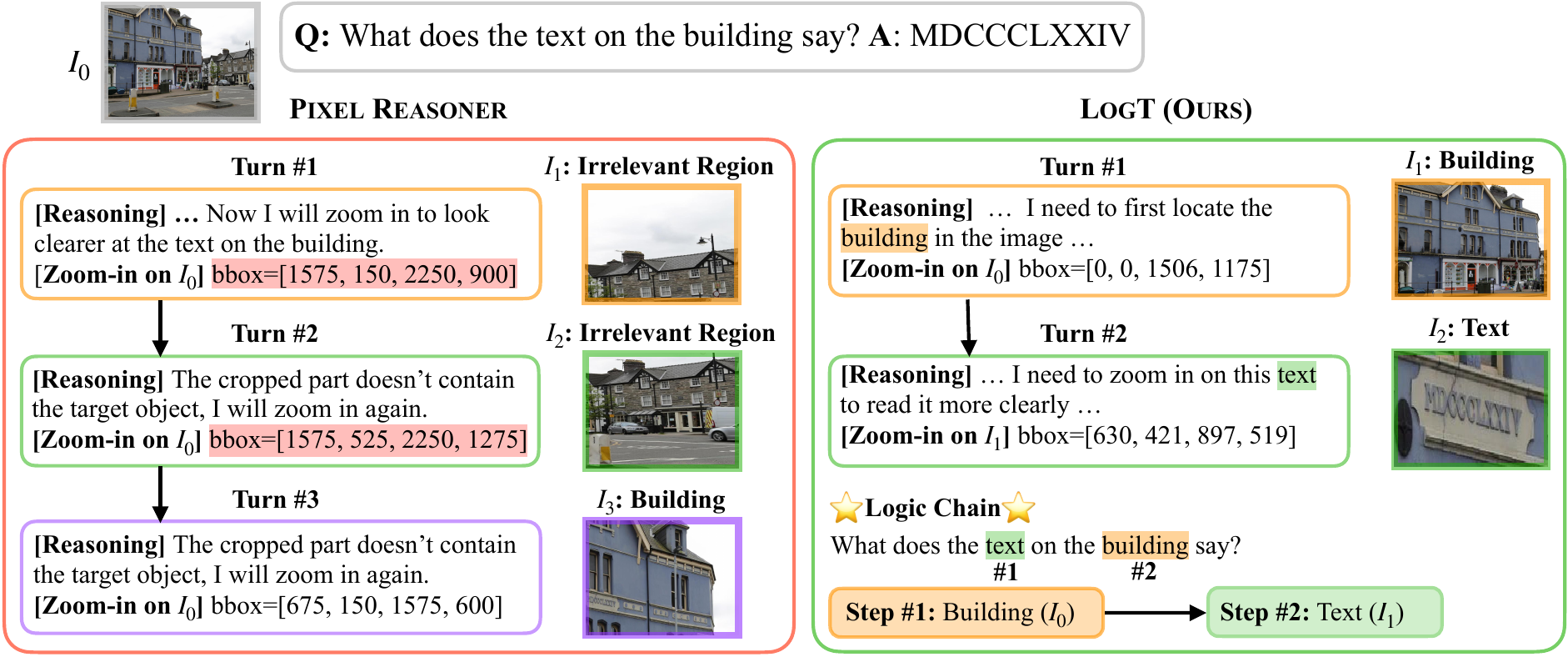}
        \caption{\textbf{\pixel vs. \ours.} The example trace in \pixel SFT data randomly introduces errors when synthesizing multi-turn tool calls, and the erroneous bounding boxes are highlighted in {\setlength{\fboxsep}{1pt}\colorbox{myred}{red}}. In contrast, ours follows a coherent logic chain: first locating the building, and then locating the text.}
        \label{fig:pixel_vs_logt}
    \end{subfigure}
    \vspace{0.5em}
    \begin{subfigure}[t]{0.75\linewidth}
        \centering
        \includegraphics[width=\linewidth]{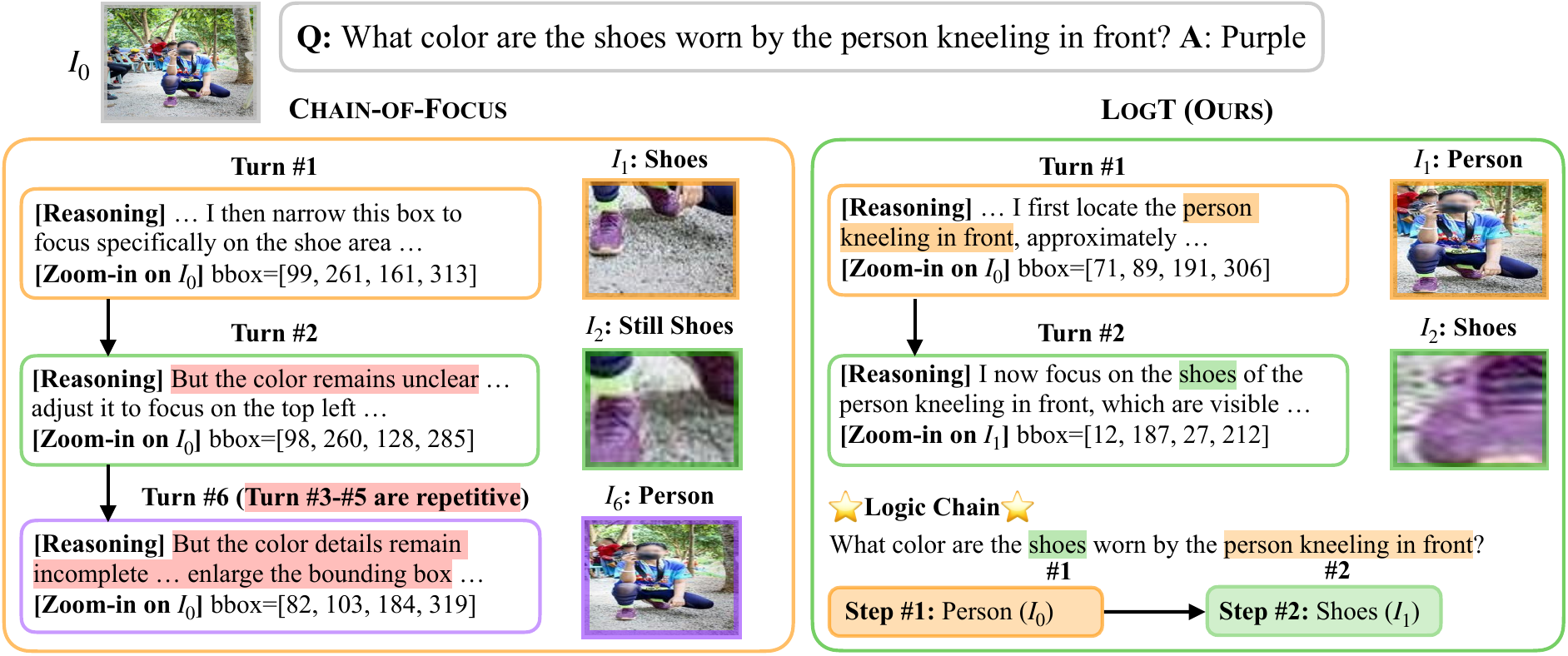}
        \caption{\textbf{\cof vs. \ours.} The example trace in \cof SFT data repeatedly zooms in on the shoes over six turns, despite correctly identifying the target region in the first turn. This redundant reasoning, highlighted in {\setlength{\fboxsep}{1pt}\colorbox{myred}{red}}, reflects tool call inefficiency. In contrast, \ours demonstrates a coherent reasoning chain: first locating the person, then progressively narrowing down to the shoes.}
        \label{fig:cof_vs_logt}
    \end{subfigure}
    \vspace{0.5em}
    \begin{subfigure}[t]{0.75\linewidth}
        \centering
        \includegraphics[width=\linewidth]{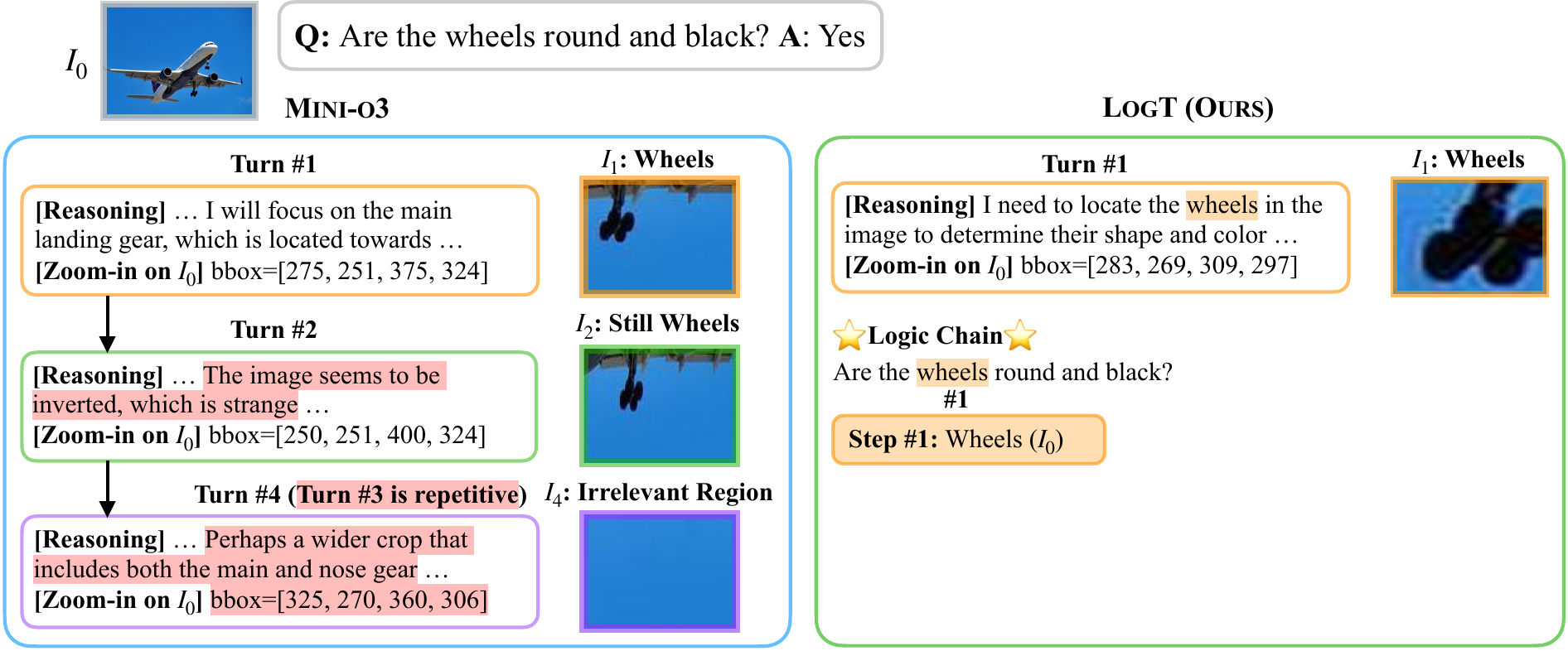}
        \caption{\textbf{\minio vs. \ours.} The example trace in \minio SFT data correctly finds the wheels in the first turn but mistakenly claims that the crop returned by \zoom is inverted, leading to two redundant \zoom calls. Ultimately, it zooms in on an entirely irrelevant region. The redundant reasoning and the incorrect \zoom call are highlighted in {\setlength{\fboxsep}{1pt}\colorbox{myred}{red}}. In contrast, \ours makes a single \zoom call to locate the wheels and then directly reaches the correct answer.}
        \label{fig:minio3_vs_logt}
    \end{subfigure}

    \caption{\textbf{Qualitative case study of SFT data synthesized by \pixel, \cof, \minio and \ours.}}
    \label{fig:sft_trace_comparison}
\end{figure*}

%% file: tables/other_baselines.tex
\begin{table}[!t]
\small
\centering
\caption{\textbf{Performance comparison between our models and additional baselines.} We extend the main results in~\cref{tab:main_results} by including additional open-source models (\internvl, \llavaov), modular-agent frameworks (\sketch, \pyvision), and visual-search-based methods (\seal, \dyfo).
The {\setlength{\fboxsep}{1pt}\colorbox{green!25}{green}} rows refer to our models.
Both \ours-SFT and \ours-RL significantly outperform these baselines across all seven test sets, further demonstrating the effectiveness of our approach.}

\begin{adjustbox}{max width=\columnwidth}
\begin{tabular}{lccccccccccc}
\toprule 
 & \multicolumn{2}{c}{Data} & \multirow{2}{*}{\vstar} & \multicolumn{3}{c}{\visualp} & \multicolumn{2}{c}{\hrbench} & \multirow{2}{*}{\begin{tabular}[c]{@{}c@{}}MME- \\ RW-Lite\end{tabular}} & \multirow{2}{*}{Avg.} \\
\cmidrule(lr){2-3}\cmidrule(lr){5-7}\cmidrule(lr){8-9}
 & SFT & RL &  & Easy & Medium & Hard & 4K & 8K &  &  \\
\midrule
\rowcolor{gray!20}
\multicolumn{11}{l}{\textit{Models with Tools}} \\
\sketch         & -   & -    & 80.30 & - & - & - & - & - & - & - \\
\pyvision       & -   & -    & 75.90 & - & - & - & - & - & - & - \\
\midrule
\rowcolor{gray!20}
\multicolumn{11}{l}{\textit{Open-sourced Models w/o Tools}} \\
\internvl & -   & -    & 60.73 & 29.79 & 10.07 & 13.21 & 57.38 & 49.00 & 41.12 & 37.33 \\
\llavaov & -   & -    & 72.25 & 41.84 & 16.42 & 14.15 & 64.62 & 58.00 & 44.71 & 44.57 \\
\midrule

\rowcolor{gray!20}
\multicolumn{11}{l}{\textit{Open-sourced Models with Visual Search Tools}} \\
\seal & -   & -    & 75.39 & - & - & - & - & - & - & - \\
\dyfo & -   & -    & 81.15 & - & - & - & - & - & - & - \\
\midrule

\rowcolor{green!5}
\ours-SFT           & 7.1k & 0     & \textbf{87.43} & \textbf{60.99} & \textbf{41.04} & \textbf{39.62} & \textbf{72.00} & \textbf{70.12} & \textbf{45.96} & \textbf{59.59} \\
\rowcolor{green!5}
\ours-RL           & 7.1k & 6.4k  & \textbf{92.15} & \textbf{70.92} & \textbf{52.24} & \textbf{51.89} & \textbf{76.12} & \textbf{74.62} & \textbf{53.00} & \textbf{67.28} \\
\bottomrule
\end{tabular}
\label{tab:other_baselines}
\end{adjustbox}
\end{table}

%% file: tables/internvl.tex
\begin{table}[!t]
\small
\centering
\caption{\textbf{Performance of fine-tuning \internvl with \ours SFT data.}
We fine-tune \internvl using 7.1k samples synthesized by \ours.
The {\setlength{\fboxsep}{1pt}\colorbox{green!25}{green}} rows refer to our models.
The resulting model outperforms the base model across all test sets, improving overall performance by 14.00 points.}

\begin{adjustbox}{max width=\columnwidth}
\begin{tabular}{lccccccccccc}
\toprule 
 & \multicolumn{2}{c}{Data} & \multirow{2}{*}{\vstar} & \multicolumn{3}{c}{\visualp} & \multicolumn{2}{c}{\hrbench} & \multirow{2}{*}{\begin{tabular}[c]{@{}c@{}}MME- \\ RW-Lite\end{tabular}} & \multirow{2}{*}{Avg.} \\
\cmidrule(lr){2-3}\cmidrule(lr){5-7}\cmidrule(lr){8-9}
 & SFT & RL &  & Easy & Medium & Hard & 4K & 8K &  &  \\
\midrule
\internvl & -   & -    & 60.73 & 29.79 & 10.07 & 13.21 & 57.38 & 49.00 & 41.12 & 37.33 \\
\midrule

\rowcolor{green!5}
+\ours-SFT           & 7.1k & 0     & \textbf{75.92} & \textbf{56.74} & \textbf{27.24} & \textbf{28.30} & \textbf{66.12} & \textbf{63.38} & \textbf{41.64} & \textbf{51.33} \\
\rowcolor{green!5}
$\Delta$ vs. Base Model & - & -     & \Inc{15.19} & \Inc{26.95} & \Inc{17.17} & \Inc{15.09} & \Inc{8.74} & \Inc{14.38} & \Inc{0.52} & \Inc{14.00} \\
\bottomrule
\end{tabular}
\label{tab:internvl}
\end{adjustbox}
\end{table}

%% file: figures/humaneval_prompt.tex
\begin{figure}[!tb]
\centering
\begin{humanevalbox}{Instructions for Human Evaluation}
\begin{lstlisting}
Which model response do you prefer?
Evaluation Criteria:
1. Answer Correctness & No Hallucination: Does the model arrive at the correct answer? It is NOT preferred that the answer is obtained by hallucination.
2. Coherent Logic: Does the model's reasoning make sense and match how you would solve the problem?
3. Tool Call Efficiency: Does the model use the minimum necessary number of tool calls (fewer iterations) to answer the question?

[Question]
...
[Ground Truth]
...
[Model A's Response]
...
[Model B's Response]
...

Your Preference:
Please select one:
Model A; Tie; Model B
\end{lstlisting}
\end{humanevalbox}
\caption{\textbf{Instructions for Human Evaluation.}}
\label{fig:humaneval_prompt}
\end{figure}

%% file: figures/humaneval_breakdown_figure.tex
\begin{figure}[!tb]
    \centering
    \includegraphics[width=0.45\linewidth]{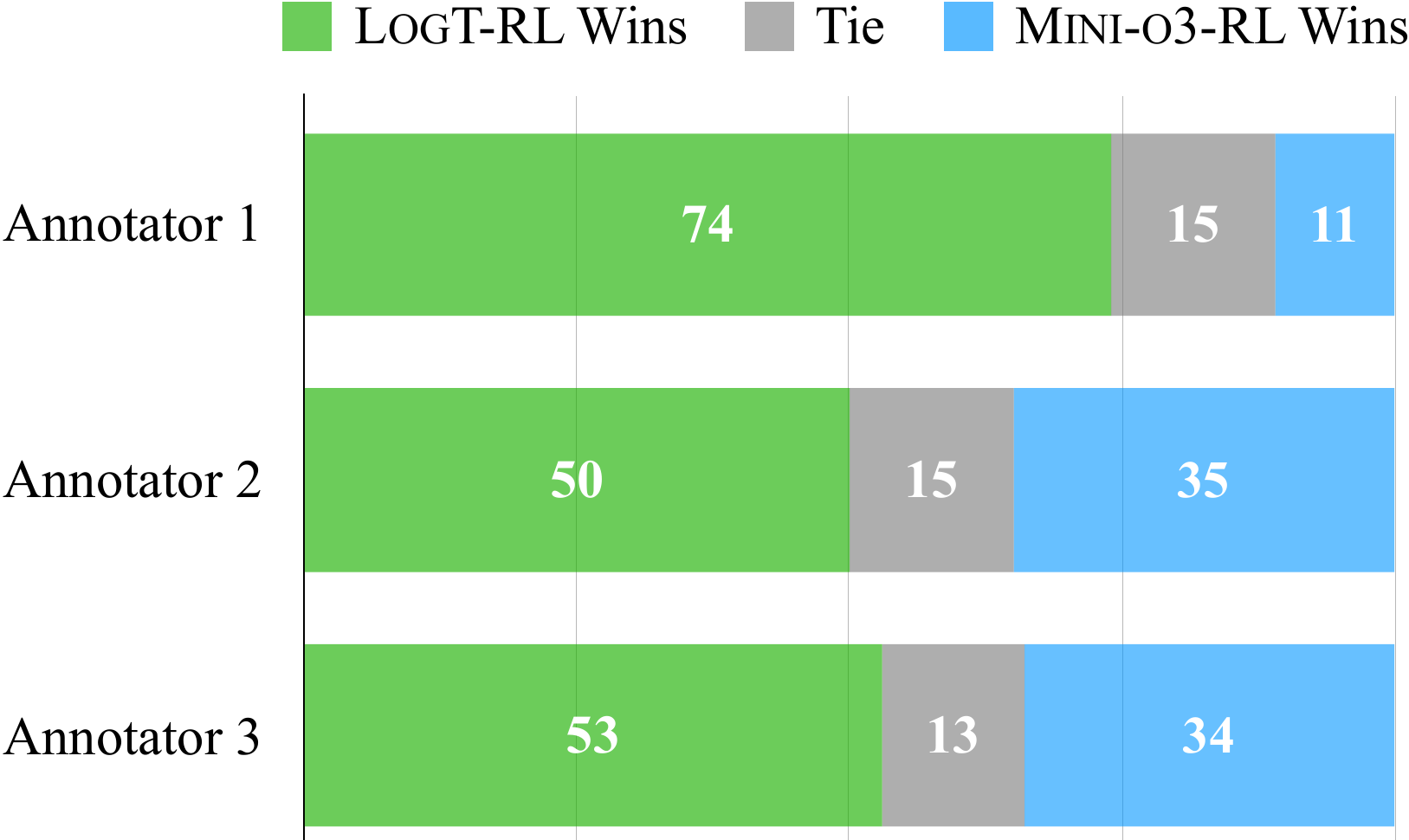}
    \caption{\textbf{Per-annotator breakdown of human evaluation.} All three annotators preferred responses generated by \ours-RL over those from \minio-RL.}
    \label{fig:huameval_breakdown}
\end{figure}

%% file: tables/ablation_qa.tex
\begin{table}[!t]
\small
\centering
\caption{\textbf{Ablation on question difficulty.} Training the VLM on the harder half of our SFT dataset leads to significantly improved performance on challenging tasks like those in \visualp-Hard.}
\label{tab:ablation_qa}
\begin{tabular}{lcccc}
\toprule
\multirow{2}{*}{Split} & \multirow{2}{*}{\vstar} & \multicolumn{3}{c}{\visualp}                  \\
\cmidrule(lr){3-5}
                       &                     & Easy           & Medium         & Hard           \\
\midrule
Easier                 & 85.34               & \textbf{59.57} & \textbf{35.07} & 35.85          \\
Harder                 & \textbf{87.43}      & 58.87          & 34.70           & \textbf{40.57} \\
\bottomrule
\end{tabular}
\end{table}

%% file: tables/ablation_chain.tex
\begin{table}[!t]
\small
\centering
\caption{
\textbf{Ablation on logic chain-driven reasoning trace synthesis.}
Incorporating logical chains into the data generation process of \ours improves overall SFT performance and yields substantial additional gains after subsequent RL.
}
\label{tab:ablation_chain}
\begin{tabular}{lccccccccc}
\hline
 &
   &
   &
  \multicolumn{3}{c}{\visualp} &
  \multicolumn{2}{c}{\hrbench} &
   &
  \multicolumn{1}{l}{} \\ \cline{4-8}
\multirow{-2}{*}{\begin{tabular}[c]{@{}l@{}}Training\\ Recipe\end{tabular}} &
  \multirow{-2}{*}{\begin{tabular}[c]{@{}c@{}}Logic\\ Chain\end{tabular}} &
  \multirow{-2}{*}{\vstar} &
  Easy &
  Medium &
  Hard &
  4K &
  8K &
  \multirow{-2}{*}{\begin{tabular}[c]{@{}c@{}}MME-\\ RW-Lite\end{tabular}} &
  \multicolumn{1}{l}{\multirow{-2}{*}{Avg.}} \\ \hline
 &
  \xsymbol &
  87.43 &
  59.57 &
  38.06 &
  39.62 &
  \textbf{72.25} &
  69.50 &
  44.09 &
  58.65 \\
\multirow{-2}{*}{SFT} &
  \csymbol &
  87.43 &
  \textbf{60.99} &
  \textbf{41.04} &
  39.62 &
  72.00 &
  \textbf{70.12} &
  \textbf{45.96} &
  \textbf{59.59} \\ \hline
 &
  \xsymbol &
  90.05 &
  66.67 &
  46.27 &
  46.23 &
  74.75 &
  \textbf{75.12} &
  51.80 &
  64.41 \\
\multirow{-2}{*}{SFT+RL} &
  \csymbol &
  \textbf{92.15} &
  \textbf{70.92} &
  \textbf{52.24} &
  \textbf{51.89} &
  \textbf{76.12} &
  74.62 &
  \textbf{53.00} &
  \textbf{67.28} \\ \hline
\end{tabular}
\end{table}

%% file: tables/ablation_models.tex
\begin{table}[!t]
\small
\centering
\caption{\textbf{Ablation on models used for data synthesis.} We replace the models used by \ours in~\cref{sec:logt} with much smaller variants and re-synthesize 4k SFT samples. After SFT, the average performance across \vstar, \visualp, and \hrbench remains nearly unchanged, indicating that our method is robust to the choice of synthesis models.}
\begin{tabular}{llc}
\toprule
Model Usage & Avg. \\
\midrule
Original  & 59.55 \\
Qwen3-VL-8B, Qwen2.5-VL-32B \& GPT-OSS-20B & 59.43 \\
\bottomrule
\end{tabular}
\label{tab:ablation_models}
\end{table}

%% file: tables/error_rate.tex
\begin{table}[!t]
    \centering
    \caption{\textbf{Error rates across different stages.}
    We randomly sampled 100 samples from each stage and manually evaluated their error rates.}
    \label{tab:error_rate}
    \begin{tabular}{lc}
        \toprule
        \textbf{Stage} & \textbf{Error Rate (\%)} \\
        \midrule
        QA Generation \& Filtering & 12 \\
        Logic Chain Generation & 2 \\
        Reasoning Trace Generation & 7 \\
        \bottomrule
    \end{tabular}
\end{table}

%% file: tables/zero_shot_rotation.tex
\begin{table}[!t]
\small
\centering
\caption{\textbf{Performance on \vstar-Rotated.} We construct \vstar-Rotated by rotating images in \vstar by angles randomly sampled from $30^{\circ}$ to $330^{\circ}$. \ours-RL uses a rotation tool at inference time that was not seen during training, improving over the base model by 21.99 points.}
\begin{tabular}{lc}
\toprule
Model & \vstar-Rotated \\
\midrule
Qwen2.5-VL-7B & 46.07 \\
\midrule
\rowcolor{green!5}
\ours-RL (with Rotation Tool) & \textbf{68.06} \\
\rowcolor{green!5}
$\Delta$ vs. Base Model & \Inc{21.99} \\
\bottomrule
\end{tabular}
\label{tab:rotation}
\end{table}

%% file: figures/eval_trace_comparison_figure.tex
\begin{figure}[!tb]
    \centering
    \includegraphics[width=0.70\linewidth]{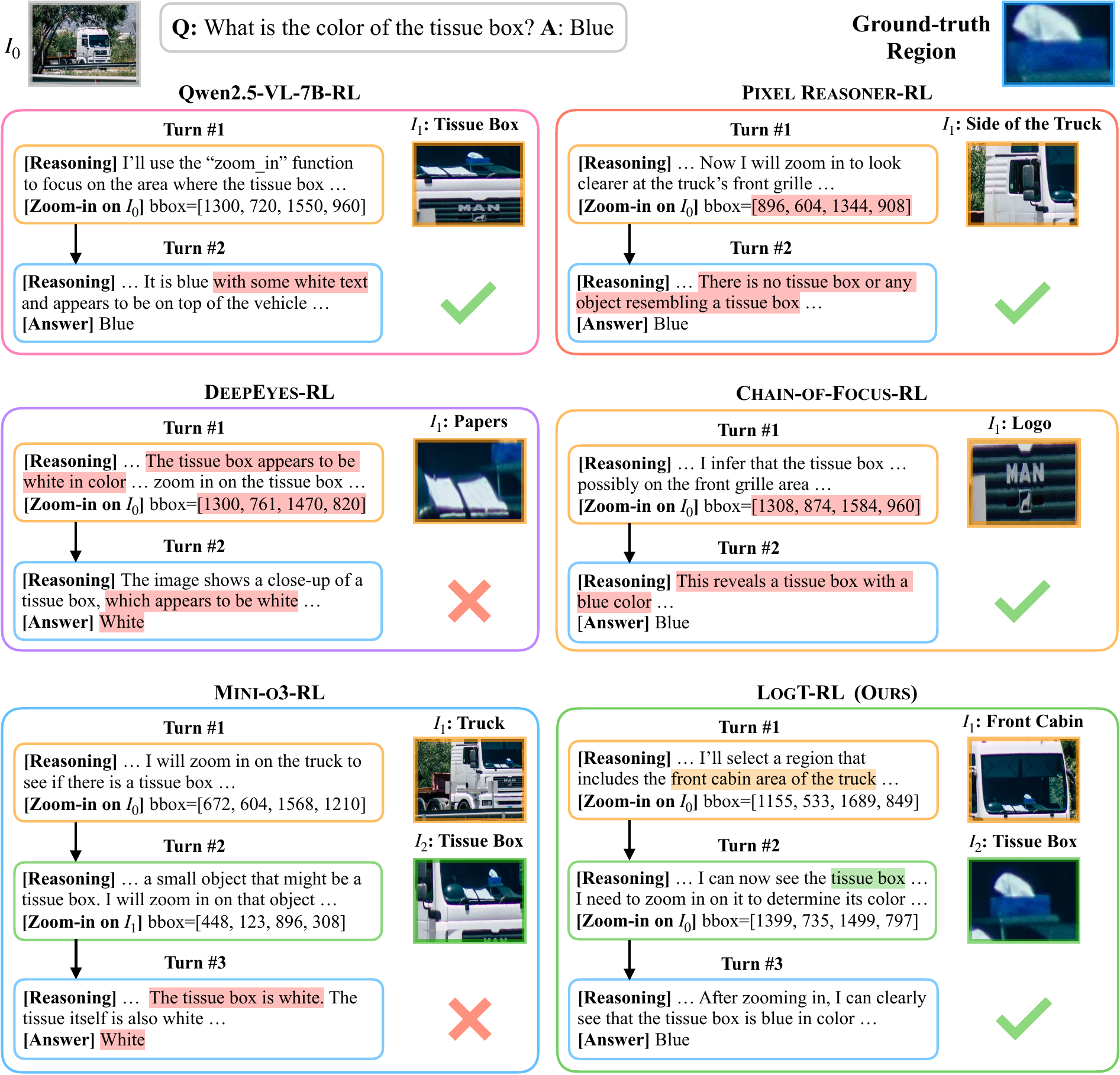}
    \caption{\textbf{Qualitative case study comparing model responses to a \vstar question between baseline models and \ours-RL.}
    We show the generated reasoning traces, including \zoom calls, for both baseline models and \ours-RL.
    While \qwenvl-RL, \pixel-RL, and \cof-RL produce the correct final answer, their responses contain hallucinations or self-contradictions. The remaining baselines make erroneous \zoom calls and ultimately produce incorrect answers.
    In contrast, \ours-RL follows a coherent and consistent reasoning chain: it first identifies the front cabin of the truck, then locates the tissue box, and finally answers the question correctly without hallucinations.}
    \label{fig:eval_trace_comparison}
\end{figure}

%% file: meta_arxiv.bbl
\begin{thebibliography}{50}
\providecommand{\natexlab}[1]{#1}
\providecommand{\url}[1]{\texttt{#1}}
\expandafter\ifx\csname urlstyle\endcsname\relax
  \providecommand{\doi}[1]{doi: #1}\else
  \providecommand{\doi}{doi: \begingroup \urlstyle{rm}\Url}\fi

\bibitem[Alayrac et~al.(2022)Alayrac, Donahue, Luc, Miech, Barr, Hasson, Lenc,
  Mensch, Millican, Reynolds, et~al.]{alayrac2022flamingo}
Jean-Baptiste Alayrac, Jeff Donahue, Pauline Luc, Antoine Miech, Iain Barr,
  Yana Hasson, Karel Lenc, Arthur Mensch, Katherine Millican, Malcolm Reynolds,
  et~al.
\newblock Flamingo: a visual language model for few-shot learning.
\newblock In \emph{Advances in Neural Information Processing Systems}, pages
  23716--23736, 2022.

\bibitem[Bai et~al.(2025{\natexlab{a}})Bai, Cai, Chen, Chen, Chen, Cheng, Deng,
  Ding, Gao, Ge, Ge, et~al.]{bai2025qwen3vltechnicalreport}
Shuai Bai, Yuxuan Cai, Ruizhe Chen, Keqin Chen, Xionghui Chen, Zesen Cheng,
  Lianghao Deng, Wei Ding, Chang Gao, Chunjiang Ge, Wenbin Ge, et~al.
\newblock Qwen3-vl technical report.
\newblock \emph{arXiv preprint arXiv:2511.21631}, 2025{\natexlab{a}}.

\bibitem[Bai et~al.(2025{\natexlab{b}})Bai, Chen, Liu, Wang, Ge, Song, Dang,
  Wang, Wang, Tang, et~al.]{bai2025qwen2}
Shuai Bai, Keqin Chen, Xuejing Liu, Jialin Wang, Wenbin Ge, Sibo Song, Kai
  Dang, Peng Wang, Shijie Wang, Jun Tang, et~al.
\newblock Qwen2.5-vl technical report.
\newblock \emph{arXiv preprint arXiv:2502.13923}, 2025{\natexlab{b}}.

\bibitem[Fu et~al.(2025)Fu, Hamman, and Dutta]{fu2025tshirt}
Yanjun Fu, Faisal Hamman, and Sanghamitra Dutta.
\newblock T-{SHIRT}: Token-selective hierarchical data selection for
  instruction tuning.
\newblock In \emph{The Thirty-ninth Annual Conference on Neural Information
  Processing Systems}, 2025.

\bibitem[Guo et~al.(2025)Guo, Zheng, Li, Bai, Li, Wang, Zhu, Neubig, Chen, and
  Yue]{guo_2025_mammoth}
Jiawei Guo, Tianyu Zheng, Yizhi Li, Yuelin Bai, Bo~Li, Yubo Wang, King Zhu,
  Graham Neubig, Wenhu Chen, and Xiang Yue.
\newblock {MA}mmo{TH}-{VL}: Eliciting multimodal reasoning with instruction
  tuning at scale.
\newblock In \emph{Proceedings of the 63rd Annual Meeting of the Association
  for Computational Linguistics (Volume 1: Long Papers)}, pages 13869--13920,
  2025.

\bibitem[Hamman et~al.(2025)Hamman, Dissanayake, Fu, and
  Dutta]{hamman2025fewshot}
Faisal Hamman, Pasan Dissanayake, Yanjun Fu, and Sanghamitra Dutta.
\newblock Few-shot knowledge distillation of {LLM}s with counterfactual
  explanations.
\newblock In \emph{The Thirty-ninth Annual Conference on Neural Information
  Processing Systems}, 2025.

\bibitem[Hu et~al.(2024)Hu, Shi, Fu, Roth, Ostendorf, Zettlemoyer, Smith, and
  Krishna]{hu2024visual}
Yushi Hu, Weijia Shi, Xingyu Fu, Dan Roth, Mari Ostendorf, Luke Zettlemoyer,
  Noah~A Smith, and Ranjay Krishna.
\newblock Visual sketchpad: Sketching as a visual chain of thought for
  multimodal language models.
\newblock In \emph{Advances in Neural Information Processing Systems}, pages
  139348--139379, 2024.

\bibitem[Huang et~al.(2025{\natexlab{a}})Huang, Xu, Zhou, Liu, Xiao, Ou, Ji,
  Li, and Yuan]{huang2025samr1}
Jiaqi Huang, Zunnan Xu, Jun Zhou, Ting Liu, Yicheng Xiao, Mingwen Ou, Bowen Ji,
  Xiu Li, and Kehong Yuan.
\newblock Sam-r1: Leveraging sam for reward feedback in multimodal segmentation
  via reinforcement learning.
\newblock \emph{arXiv preprint arXiv: 2505.22596}, 2025{\natexlab{a}}.

\bibitem[Huang et~al.(2025{\natexlab{b}})Huang, Ji, Rajan, Cai, Xiao, Wang, Hu,
  and Lee]{huang2025vista}
Zeyi Huang, Yuyang Ji, Anirudh~Sundara Rajan, Zefan Cai, Wen Xiao, Haohan Wang,
  Junjie Hu, and Yong~Jae Lee.
\newblock Visualtoolagent (vista): A reinforcement learning framework for
  visual tool selection.
\newblock \emph{arXiv preprint arXiv: 2505.20289}, 2025{\natexlab{b}}.

\bibitem[Hurst et~al.(2024)Hurst, Lerer, Goucher, Perelman, Ramesh, Clark,
  Ostrow, Welihinda, Hayes, Radford, et~al.]{hurst2024gpt}
Aaron Hurst, Adam Lerer, Adam~P Goucher, Adam Perelman, Aditya Ramesh, Aidan
  Clark, AJ~Ostrow, Akila Welihinda, Alan Hayes, Alec Radford, et~al.
\newblock Gpt-4o system card.
\newblock \emph{arXiv preprint arXiv:2410.21276}, 2024.

\bibitem[Kirillov et~al.(2023)Kirillov, Mintun, Ravi, Mao, Rolland, Gustafson,
  Xiao, Whitehead, Berg, Lo, et~al.]{kirillov2023segment}
Alexander Kirillov, Eric Mintun, Nikhila Ravi, Hanzi Mao, Chloe Rolland, Laura
  Gustafson, Tete Xiao, Spencer Whitehead, Alexander~C Berg, Wan-Yen Lo, et~al.
\newblock Segment anything.
\newblock In \emph{Proceedings of the IEEE/CVF International Conference on
  Computer Vision (ICCV)}, pages 4015--4026, 2023.

\bibitem[Lai et~al.(2025)Lai, Li, Li, Liu, Li, and Zhao]{lai2025mini}
Xin Lai, Junyi Li, Wei Li, Tao Liu, Tianjian Li, and Hengshuang Zhao.
\newblock Mini-o3: Scaling up reasoning patterns and interaction turns for
  visual search.
\newblock \emph{arXiv preprint arXiv:2509.07969}, 2025.

\bibitem[Li et~al.(2025{\natexlab{a}})Li, Wang, Fu, Yue, Cai, Zhu, Liu, Guo,
  Neiswanger, Huang, Goldstein, and Goldblum]{li2025zebracot}
Ang Li, Charles Wang, Deqing Fu, Kaiyu Yue, Zikui Cai, Wang~Bill Zhu, Ollie
  Liu, Peng Guo, Willie Neiswanger, Furong Huang, Tom Goldstein, and Micah
  Goldblum.
\newblock Zebra-cot: A dataset for interleaved vision language reasoning.
\newblock \emph{arXiv preprint arXiv:2507.16746}, 2025{\natexlab{a}}.

\bibitem[Li et~al.(2024)Li, Zhang, Guo, Zhang, Li, Zhang, Zhang, Zhang, Li,
  Liu, and Li]{li2024llavaonevision}
Bo~Li, Yuanhan Zhang, Dong Guo, Renrui Zhang, Feng Li, Hao Zhang, Kaichen
  Zhang, Peiyuan Zhang, Yanwei Li, Ziwei Liu, and Chunyuan Li.
\newblock Llava-onevision: Easy visual task transfer.
\newblock \emph{arXiv preprint arXiv:2408.03326}, 2024.

\bibitem[Li et~al.(2025{\natexlab{b}})Li, Xu, Zhao, and Peng]{li2025dyfo}
Geng Li, Jinglin Xu, Yunzhen Zhao, and Yuxin Peng.
\newblock Dyfo: A training-free dynamic focus visual search for enhancing lmms
  in fine-grained visual understanding.
\newblock In \emph{Proceedings of the IEEE Conference on Computer Vision and
  Pattern Recognition (CVPR)}, pages 9098--9108, 2025{\natexlab{b}}.

\bibitem[Li et~al.(2026)Li, Tu, Su, Liu, Huang, Alinejad-Rokny, Wong, Lin, and
  Yang]{li2026toolrm}
Renhao Li, Jianhong Tu, Yang Su, Yantao Liu, Fei Huang, Hamid Alinejad-Rokny,
  Derek~F. Wong, Junyang Lin, and Min Yang.
\newblock Toolrm: Towards agentic tool-use reward modeling.
\newblock \emph{arXiv preprint arXiv:2510.26167}, 2026.

\bibitem[Li et~al.(2025{\natexlab{c}})Li, Li, Gao, Pi, Hu, and
  Zhang]{li2025looklessreasonmore}
Xuchen Li, Xuzhao Li, Jiahui Gao, Renjie Pi, Shiyu Hu, and Wentao Zhang.
\newblock Look less, reason more: Rollout-guided adaptive pixel-space
  reasoning.
\newblock \emph{arXiv preprint arXiv:2510.01681}, 2025{\natexlab{c}}.

\bibitem[Liu et~al.(2023)Liu, Li, Wu, and Lee]{liu2023visual}
Haotian Liu, Chunyuan Li, Qingyang Wu, and Yong~Jae Lee.
\newblock Visual instruction tuning.
\newblock In \emph{Advances in Neural Information Processing Systems}, pages
  34892--34916, 2023.

\bibitem[Liu et~al.(2025)Liu, Pan, She, Gao, and Xia]{liu2025faithfulness}
Zujing Liu, Junwen Pan, Qi~She, Yuan Gao, and Guisong Xia.
\newblock On the faithfulness of visual thinking: Measurement and enhancement.
\newblock \emph{arXiv preprint arXiv:2510.23482}, 2025.

\bibitem[Lu et~al.(2022)Lu, Mishra, Xia, Qiu, Chang, Zhu, Tafjord, Clark, and
  Kalyan]{lu2022learn}
Pan Lu, Swaroop Mishra, Tony Xia, Liang Qiu, Kai-Wei Chang, Song-Chun Zhu,
  Oyvind Tafjord, Peter Clark, and Ashwin Kalyan.
\newblock Learn to explain: Multimodal reasoning via thought chains for science
  question answering.
\newblock In \emph{Advances in Neural Information Processing Systems}, pages
  2507--2521, 2022.

\bibitem[Mu et~al.(2023)Mu, Zhang, Hu, Wang, Ding, Jin, Wang, Dai, Qiao, and
  Luo]{mu2023embodied}
Yao Mu, Qinglong Zhang, Mengkang Hu, Wenhai Wang, Mingyu Ding, Jun Jin, Bin
  Wang, Jifeng Dai, Yu~Qiao, and Ping Luo.
\newblock Embodiedgpt: vision-language pre-training via embodied chain of
  thought.
\newblock In \emph{Advances in Neural Information Processing Systems}, pages
  25081--25094, 2023.

\bibitem[OpenAI(2025{\natexlab{a}})]{openai2025gptoss}
OpenAI.
\newblock gpt-oss-120b \& gpt-oss-20b model card.
\newblock \emph{arXiv preprint arXiv:2508.10925}, 2025{\natexlab{a}}.

\bibitem[OpenAI(2025{\natexlab{b}})]{openai2025thinking}
OpenAI.
\newblock Thinking with images, 2025{\natexlab{b}}.
\newblock \url{https://openai.com/index/thinking-with-images/}.

\bibitem[OpenAI(2025{\natexlab{c}})]{openaio32025}
OpenAI.
\newblock Openai o3, 2025{\natexlab{c}}.
\newblock \url{https://openai.com/index/introducing-o3-and-o4-mini/}.

\bibitem[OpenAI(2026)]{openai2026why}
OpenAI.
\newblock Why swe-bench verified no longer measures frontier coding
  capabilities, 2026.
\newblock
  \url{https://openai.com/index/why-we-no-longer-evaluate-swe-bench-verified/}.

\bibitem[Rao et~al.(2002)Rao, Zelinsky, Hayhoe, and Ballard]{rao2002eye}
Rajesh~P.N. Rao, Gregory~J. Zelinsky, Mary~M. Hayhoe, and Dana~H. Ballard.
\newblock Eye movements in iconic visual search.
\newblock \emph{Vision Research}, 42\penalty0 (11):\penalty0 1447--1463, 2002.

\bibitem[Shao et~al.(2024{\natexlab{a}})Shao, Qian, Xiao, Song, Zong, Wang,
  Liu, and Li]{shao2024visualcot}
Hao Shao, Shengju Qian, Han Xiao, Guanglu Song, Zhuofan Zong, Letian Wang,
  Yu~Liu, and Hongsheng Li.
\newblock Visual cot: Advancing multi-modal language models with a
  comprehensive dataset and benchmark for chain-of-thought reasoning.
\newblock In \emph{Advances in Neural Information Processing Systems Datasets
  and Benchmarks Track}, pages 8612--8642, 2024{\natexlab{a}}.

\bibitem[Shao et~al.(2024{\natexlab{b}})Shao, Wang, Zhu, Xu, Song, Bi, Zhang,
  Zhang, Li, Wu, and Guo]{shao2024deepseekmathpushinglimitsmathematical}
Zhihong Shao, Peiyi Wang, Qihao Zhu, Runxin Xu, Junxiao Song, Xiao Bi, Haowei
  Zhang, Mingchuan Zhang, Y.~K. Li, Y.~Wu, and Daya Guo.
\newblock Deepseekmath: Pushing the limits of mathematical reasoning in open
  language models.
\newblock \emph{arXiv preprint arXiv:2402.03300}, 2024{\natexlab{b}}.

\bibitem[Sheng et~al.(2024)Sheng, Zhang, Ye, Wu, Zhang, Zhang, Peng, Lin, and
  Wu]{sheng2024hybridflow}
Guangming Sheng, Chi Zhang, Zilingfeng Ye, Xibin Wu, Wang Zhang, Ru~Zhang,
  Yanghua Peng, Haibin Lin, and Chuan Wu.
\newblock Hybridflow: A flexible and efficient rlhf framework.
\newblock \emph{arXiv preprint arXiv: 2409.19256}, 2024.

\bibitem[Shi et~al.(2024)Shi, Ma, Liang, Diao, Ma, and
  Vosoughi]{shi2024judging}
Lin Shi, Chiyu Ma, Wenhua Liang, Xingjian Diao, Weicheng Ma, and Soroush
  Vosoughi.
\newblock Judging the judges: A systematic study of position bias in
  llm-as-a-judge.
\newblock \emph{arXiv preprint arXiv:2406.07791}, 2024.

\bibitem[Su et~al.(2025{\natexlab{a}})Su, Wang, Ren, Lin, and
  Chen]{su2025pixel}
Alex Su, Haozhe Wang, Weiming Ren, Fangzhen Lin, and Wenhu Chen.
\newblock Pixel reasoner: Incentivizing pixel space reasoning via
  curiosity-driven reinforcement learning.
\newblock In \emph{The Thirty-ninth Annual Conference on Neural Information
  Processing Systems}, 2025{\natexlab{a}}.

\bibitem[Su et~al.(2025{\natexlab{b}})Su, Li, Song, Hao, Yang, Zhang, Chen, Gu,
  Li, Qu, et~al.]{su2025openthinkimg}
Zhaochen Su, Linjie Li, Mingyang Song, Yunzhuo Hao, Zhengyuan Yang, Jun Zhang,
  Guanjie Chen, Jiawei Gu, Juntao Li, Xiaoye Qu, et~al.
\newblock Openthinkimg: Learning to think with images via visual tool
  reinforcement learning.
\newblock \emph{arXiv preprint arXiv:2505.08617}, 2025{\natexlab{b}}.

\bibitem[Sun et~al.(2025)Sun, Liang, Wei, Yu, Li, Yang, Zhou, and
  Zhang]{sun_2025_mmverify}
Linzhuang Sun, Hao Liang, Jingxuan Wei, Bihui Yu, Tianpeng Li, Fan Yang, Zenan
  Zhou, and Wentao Zhang.
\newblock {MM}-verify: Enhancing multimodal reasoning with chain-of-thought
  verification.
\newblock In \emph{Proceedings of the 63rd Annual Meeting of the Association
  for Computational Linguistics (Volume 1: Long Papers)}, pages 14100--14115,
  2025.

\bibitem[Treisman and Gelade(1980)]{treisman1980feature}
Anne~M. Treisman and Garry Gelade.
\newblock A feature-integration theory of attention.
\newblock \emph{Cognitive Psychology}, 12\penalty0 (1):\penalty0 97--136, 1980.

\bibitem[Vendrow et~al.(2025)Vendrow, Vendrow, Beery, and
  Madry]{vendrow2025largelanguagemodelbenchmarks}
Joshua Vendrow, Edward Vendrow, Sara Beery, and Aleksander Madry.
\newblock Do large language model benchmarks test reliability?
\newblock \emph{arXiv preprint arXiv:2502.03461}, 2025.

\bibitem[Wang et~al.(2025{\natexlab{a}})Wang, Gao, Gu, Pu, Cui, Wei, Liu, Jing,
  Ye, Shao, et~al.]{wang2025internvl3_5}
Weiyun Wang, Zhangwei Gao, Lixin Gu, Hengjun Pu, Long Cui, Xingguang Wei,
  Zhaoyang Liu, Linglin Jing, Shenglong Ye, Jie Shao, et~al.
\newblock Internvl3.5: Advancing open-source multimodal models in versatility,
  reasoning, and efficiency.
\newblock \emph{arXiv preprint arXiv:2508.18265}, 2025{\natexlab{a}}.

\bibitem[Wang et~al.(2025{\natexlab{b}})Wang, Ding, Zeng, Zhou, Shen, Luo, Yu,
  and Tao]{hrbench}
Wenbin Wang, Liang Ding, Minyan Zeng, Xiabin Zhou, Li~Shen, Yong Luo, Wei Yu,
  and Dacheng Tao.
\newblock Divide, conquer and combine: A training-free framework for
  high-resolution image perception in multimodal large language models.
\newblock In \emph{Proceedings of the AAAI Conference on Artificial
  Intelligence}, pages 7907--7915, 2025{\natexlab{b}}.

\bibitem[Wang et~al.(2018)Wang, Xiao, Jiang, Shao, Sun, and
  Shen]{wang18repulsion}
Xinlong Wang, Tete Xiao, Yuning Jiang, Shuai Shao, Jian Sun, and Chunhua Shen.
\newblock Repulsion loss: Detecting pedestrians in a crowd.
\newblock In \emph{Proceedings of the IEEE Conference on Computer Vision and
  Pattern Recognition (CVPR)}, pages 7774--7783, 2018.

\bibitem[Wei et~al.(2022)Wei, Wang, Schuurmans, Bosma, ichter, Xia, Chi, Le,
  and Zhou]{wei2022cot}
Jason Wei, Xuezhi Wang, Dale Schuurmans, Maarten Bosma, brian ichter, Fei Xia,
  Ed~Chi, Quoc~V Le, and Denny Zhou.
\newblock Chain-of-thought prompting elicits reasoning in large language
  models.
\newblock In \emph{Advances in Neural Information Processing Systems}, pages
  24824--24837, 2022.

\bibitem[Wolfe and Horowitz(2017)]{wolfe2017five}
Jeremy~M Wolfe and Todd~S Horowitz.
\newblock Five factors that guide attention in visual search.
\newblock \emph{Nature Human Behaviour}, 1\penalty0 (3):\penalty0 0058, 2017.

\bibitem[Wu et~al.(2025)Wu, Yang, Jiang, Li, Yan, Yu, Zhang, Zhai, and
  Nahrstedt]{wu2025vtoolr1}
Mingyuan Wu, Jingcheng Yang, Jize Jiang, Meitang Li, Kaizhuo Yan, Hanchao Yu,
  Minjia Zhang, Chengxiang Zhai, and Klara Nahrstedt.
\newblock Vtool-r1: Vlms learn to think with images via reinforcement learning
  on multimodal tool use.
\newblock \emph{arXiv preprint arXiv: 2505.19255}, 2025.

\bibitem[Wu and Xie(2024)]{wu2023vstar}
Penghao Wu and Saining Xie.
\newblock V*: Guided visual search as a core mechanism in multimodal llms.
\newblock In \emph{Proceedings of the IEEE Conference on Computer Vision and
  Pattern Recognition (CVPR)}, pages 13084--13094, 2024.

\bibitem[Xu et~al.(2025)Xu, Jin, Wu, Li, Song, Sun, and
  Yuan]{xu2025llavacotletvisionlanguage}
Guowei Xu, Peng Jin, Ziang Wu, Hao Li, Yibing Song, Lichao Sun, and Li~Yuan.
\newblock Llava-cot: Let vision language models reason step-by-step.
\newblock \emph{arXiv preprint arXiv:2411.10440}, 2025.

\bibitem[Yu et~al.(2025)Yu, Yang, Zou, Yan, and Wang]{yu2025demystify}
Zhaochen Yu, Ling Yang, Jiaru Zou, Shuicheng Yan, and Mengdi Wang.
\newblock Demystifying reinforcement learning in agentic reasoning.
\newblock \emph{arXiv preprint arXiv:2510.11701}, 2025.

\bibitem[Zhang et~al.(2025{\natexlab{a}})Zhang, Zhang, Li, Zhang, Sun, Gan,
  Yang, Pang, and Yang]{zhang2025improve}
Ruohong Zhang, Bowen Zhang, Yanghao Li, Haotian Zhang, Zhiqing Sun, Zhe Gan,
  Yinfei Yang, Ruoming Pang, and Yiming Yang.
\newblock Improve vision language model chain-of-thought reasoning.
\newblock In \emph{Proceedings of the 63rd Annual Meeting of the Association
  for Computational Linguistics (Volume 1: Long Papers)}, pages 1631--1662,
  2025{\natexlab{a}}.

\bibitem[Zhang et~al.(2025{\natexlab{b}})Zhang, Gao, Zhang, Li, Zhang, Liu,
  Yuan, Wu, Jia, Zhu, and Li]{zhang2025chain}
Xintong Zhang, Zhi Gao, Bofei Zhang, Pengxiang Li, Xiaowen Zhang, Yang Liu, Tao
  Yuan, Yuwei Wu, Yunde Jia, Song-Chun Zhu, and Qing Li.
\newblock Adaptive chain-of-focus reasoning via dynamic visual search and
  zooming for efficient vlms.
\newblock \emph{arXiv preprint arXiv:2505.15436}, 2025{\natexlab{b}}.

\bibitem[Zhang et~al.(2024)Zhang, Zhang, Tian, Fu, Zhang, Wu, Li, Wang, Wen,
  Zhang, et~al.]{zhang2024mme}
Yi-Fan Zhang, Huanyu Zhang, Haochen Tian, Chaoyou Fu, Shuangqing Zhang, Junfei
  Wu, Feng Li, Kun Wang, Qingsong Wen, Zhang Zhang, et~al.
\newblock Mme-realworld: Could your multimodal llm challenge high-resolution
  real-world scenarios that are difficult for humans?
\newblock \emph{arXiv preprint arXiv:2408.13257}, 2024.

\bibitem[Zhang et~al.(2025{\natexlab{c}})Zhang, Lu, Yin, Fu, Chen, Hu, Wen,
  Jiang, Liu, Zhang, et~al.]{zhang2025thyme}
Yi-Fan Zhang, Xingyu Lu, Shukang Yin, Chaoyou Fu, Wei Chen, Xiao Hu, Bin Wen,
  Kaiyu Jiang, Changyi Liu, Tianke Zhang, et~al.
\newblock Thyme: Think beyond images.
\newblock \emph{arXiv preprint arXiv:2508.11630}, 2025{\natexlab{c}}.

\bibitem[Zhao et~al.(2025)Zhao, Zhang, Lin, Li, Wu, Zhang, and
  Wei]{zhao2025pyvision}
Shitian Zhao, Haoquan Zhang, Shaoheng Lin, Ming Li, Qilong Wu, Kaipeng Zhang,
  and Chen Wei.
\newblock Pyvision: Agentic vision with dynamic tooling.
\newblock \emph{arXiv preprint arXiv:2507.07998}, 2025.

\bibitem[Zheng et~al.(2025)Zheng, Yang, Hong, Zhao, Xu, Yang, Shen, and
  Yu]{zheng2025deepeyes}
Ziwei Zheng, Michael Yang, Jack Hong, Chenxiao Zhao, Guohai Xu, Le~Yang, Chao
  Shen, and Xing Yu.
\newblock Deepeyes: Incentivizing ``thinking with images'' via reinforcement
  learning.
\newblock \emph{arXiv preprint arXiv:2505.14362}, 2025.

\end{thebibliography}
